\documentclass{article}

\usepackage{amsmath}
\usepackage{graphicx}

 \usepackage[preprint]{neurips_2026}

\usepackage[utf8]{inputenc} %
\usepackage[T1]{fontenc}    %
\usepackage{hyperref}       %
\usepackage{url}            %
\usepackage{booktabs}       %
\usepackage{amsfonts}       %
\usepackage{nicefrac}       %
\usepackage{microtype}      %
\usepackage{xcolor}         %
\usepackage{algorithm}
\usepackage{algpseudocode}
\usepackage{multirow}
\usepackage{tcolorbox}
\tcbuselibrary{skins,breakable}
\usepackage{caption}
\title{MotionVLA: Vision-Language-Action Model for Humanoid Motion}

\author{
Nonghai Zhang$^{1*}$\quad
Siyu Zhai$^{1*}$\quad
Yanjun Li$^{1*}$\quad
Zeyu Zhang$^{1*\dag}$\\
\textbf{Zhihan Yin}$^{1}$\quad
\textbf{Yandong Guo}$^{2}$\quad
\textbf{Boxin Shi}$^{1}$\quad
\textbf{Hao Tang}$^{1\ddag}$\\ 
[0.3em]
$^1$ School of Computer Science, Peking University\quad
$^2$ AI$^2$ Robotics\\
[0.1em]
\footnotesize $^*$Equal contribution.
$^\dag$Project lead.
$^\ddag$Corresponding author: bjdxtanghao@gmail.com.
}

\begin{document}

\maketitle

\begin{abstract}
Generating realistic humanoid motion from scene images and text involves both low-frequency pose semantics and high-frequency physical dynamics. 
However, many existing methods tokenize motion with a single shared codebook, forcing heterogeneous motion signals into the same quantization space. 
Our frequency-domain analysis of human motion data reveals a clear mismatch between single-codebook quantization and motion statistics: five DCT coefficients capture 93\% of joint-position energy but only 37\% of joint-velocity energy, which can bias quantization toward pose statistics and under-represent high-frequency velocity components. 
A second challenge lies in adapting a standard autoregressive model to effectively model high-frequency physical signals in motion sequences. Therefore, we propose \textbf{DSFT}, a dual-stream frequency tokenizer that separates motion into Base and physical streams and compresses them independently with DCT truncation and BPE. 
Furthermore, we present \textbf{MotionVLA}, a Qwen3.5-based model that arranges Base and physical tokens in a unified sequence, where Phys tokens are predicted after Base tokens. 
Experiments on HumanML3D and MBench show that, despite using a lightweight 2B backbone, MotionVLA reduces the Diversity gap to real data by over 50\% on HumanML3D and improves Motion-Condition Consistency by 3.8\% on MBench, supporting frequency-aware dual-stream decoupling as an effective formulation for autoregressive motion generation.
Code:~\url{https://github.com/AIGeeksGroup/MotionVLA}.
Website:~\url{https://aigeeksgroup.github.io/MotionVLA}.
\end{abstract}

\section{Introduction}

\begin{figure}[t]
  \centering
\includegraphics[width=\linewidth]{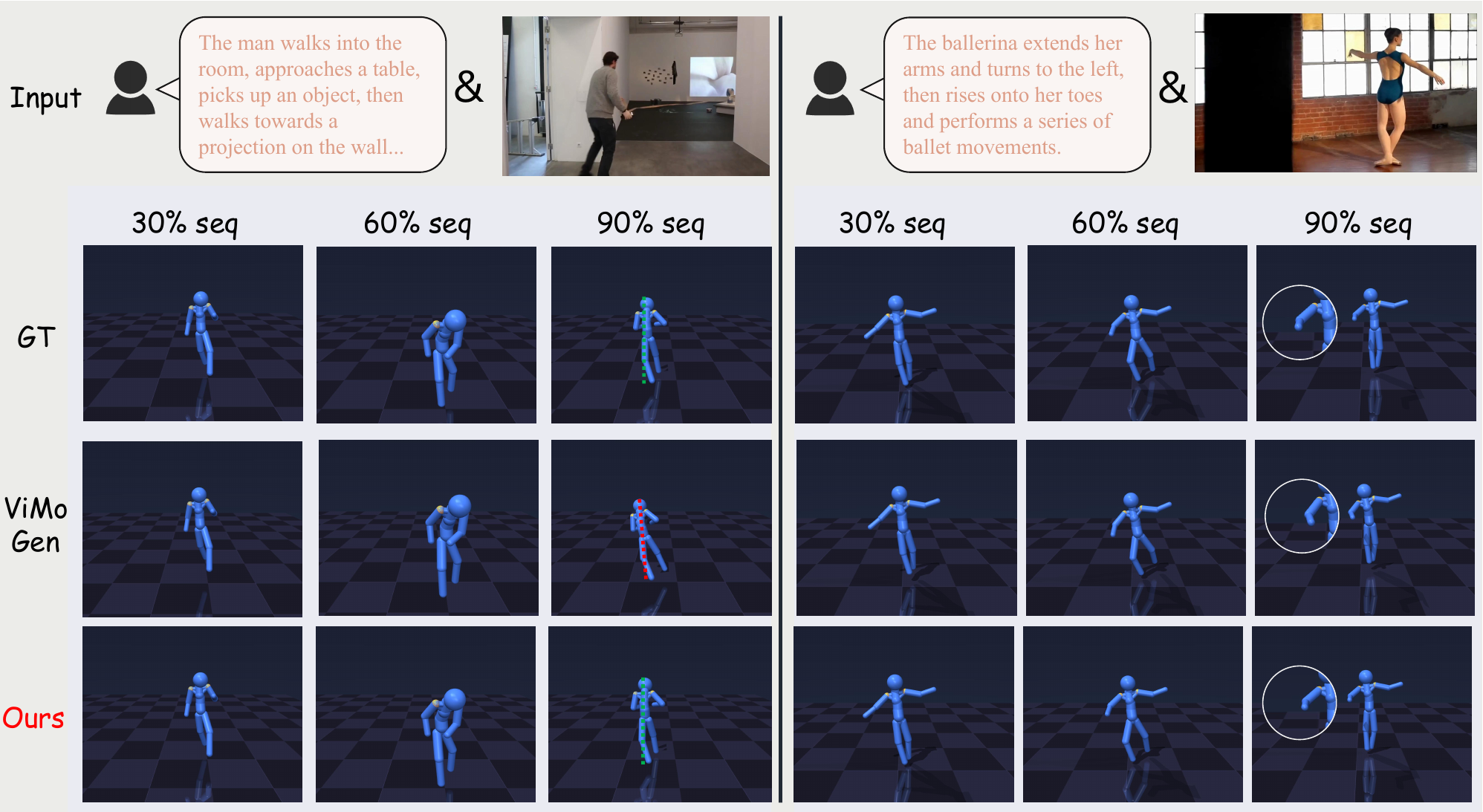}
  \caption{Given a text description and a scene video as input, MotionVLA generates
           motions that closely track the ground truth (GT) across the full sequence
           (frames at 30\%, 60\%, and 90\% shown).
           ViMoGen~\cite{vimogen}, which relies on a single-stream tokenizer,
           exhibits temporal drift and joint instability that accumulate
           over time (highlighted in white circles).}
  \label{fig:intro}
  \vspace{-0.4cm}
\end{figure}

Fine-grained humanoid motion generation is a core capability for embodied intelligence, character animation, and scene-aware action synthesis. In recent years, progress in this area has largely followed an autoregressive paradigm, in which motion is discretized into token sequences and generated step by step with Transformer-based models~\cite{t2m_gpt, motiongpt, momask, mg_motionllm, genm3, mosa, vaswani2017attention}. At the same time, recent advances in vision-language-action (VLA) modeling~\cite{openvla, pi0} highlight the value of grounding action generation in scene observations and language instructions, thereby motivating scene-aware vision-language-to-motion generation~\cite{vimogen}.

However, although recent studies have shown that applying DCT before quantization can bring clear advantages to long-horizon autoregressive generation~\cite{fast, lg_tok, motok}, these tokenizers still encode motion within a unified tokenization space, implicitly treating heterogeneous motion components as if they followed similar statistics. Our analysis shows that this assumption does not hold well for human motion: joint positions are strongly low-frequency, with the first five DCT coefficients covering 93\% of their energy, whereas joint velocities are markedly high-frequency, with the same five coefficients covering only 37\%. Consequently, such a shared tokenization mechanism is naturally biased toward low-frequency pose structure, while high-frequency physical signals are more easily under-represented.

At the same time, this representation issue directly creates a second challenge for autoregressive generation. Because a standard autoregressive model predicts motion tokens from a unified codebook, it is naturally encouraged to model the dominant low-frequency pose structure first, while the weaker high-frequency physical signals are less reliably preserved. As generation proceeds over time, this imbalance makes it difficult for the model to faithfully maintain fine-grained physical dynamics, causing errors in contact and motion stability to accumulate. In practice, this often manifests as artifacts such as temporal drift, foot sliding, and contact distortion~\cite{discord}, especially in long-horizon generation~\cite{discord, lmr}.

However, all existing tokenization methods, including frequency-domain ones, share an unresolved structural limitation: each frame of motion is represented by a single discrete token, forcing signals with fundamentally different frequency profiles into a single codebook.
We quantify this imbalance directly on HumanML3D~\cite{humanml3d}.
Joint positions are dominated by low frequencies: five DCT coefficients reconstruct 93\% of position energy.
Joint velocities, as first derivatives of positions, obey the differentiation theorem, which scales each DCT coefficient by its frequency index and inherently amplifies high-frequency components.
The same five coefficients capture only 37\% of velocity energy, a gap exceeding fifty percentage points.
When BPE is applied to a single concatenated feature, low-frequency position statistics dominate the vocabulary, causing the codebook to effectively discard the high-frequency velocity signal as noise.
The practical consequence is visible in Figure~\ref{fig:intro}: motions generated with a single-stream tokenizer accumulate drift and joint instability over time, whereas our method tracks the ground truth throughout the sequence.
Prior work has documented these artifacts, including foot sliding and contact distortion~\cite{discord}, and proposed post-hoc corrections at the decoding stage~\cite{discord, lmr}, but no prior method removes the structural cause.

To address these challenges, we propose MotionVLA, a vision-language-to-motion framework built on a dual-stream representation of human motion. Its core component, DSFT, separates motion into a Base stream that captures joint-position semantics and a physical stream that captures joint-velocity dynamics, and tokenizes them independently in the frequency domain. Building on this representation, MotionVLA arranges the two streams in a unified autoregressive sequence, where Phys tokens are generated after Base tokens so that physical-signal prediction can leverage the preceding pose context through causal attention. As show in Figure ~\ref{fig:intro} our method explicitly decouples low-frequency semantic structure from high-frequency physical dynamics while preserving a simple autoregressive formulation.
Our contributions are threefold. 
\begin{itemize}
    \item we propose \textbf{DSFT}, a dual-stream tokenizer that separates motion into Base and Phys streams and compresses them independently in the frequency domain, addressing the mismatch between unified tokenization and heterogeneous motion statistics. 
    
    \item We present \textbf{MotionVLA}, a vision-language-to-motion framework that models human motion generation as a unified autoregressive process over decoupled semantic and physical token streams. 

    \item Experiments on HumanML3D and MBench show that our method reduces the Diversity gap to real data by over 50\% on HumanML3D, while improving Motion-Condition Consistency from 0.53 to 0.55 and reducing Foot Sliding from 0.0051 to 0.0049 on MBench.
\end{itemize}

\section{The Proposed Method}

\subsection{Overview}
\begin{figure*}[!t]
  \centering
  \includegraphics[width=\linewidth]{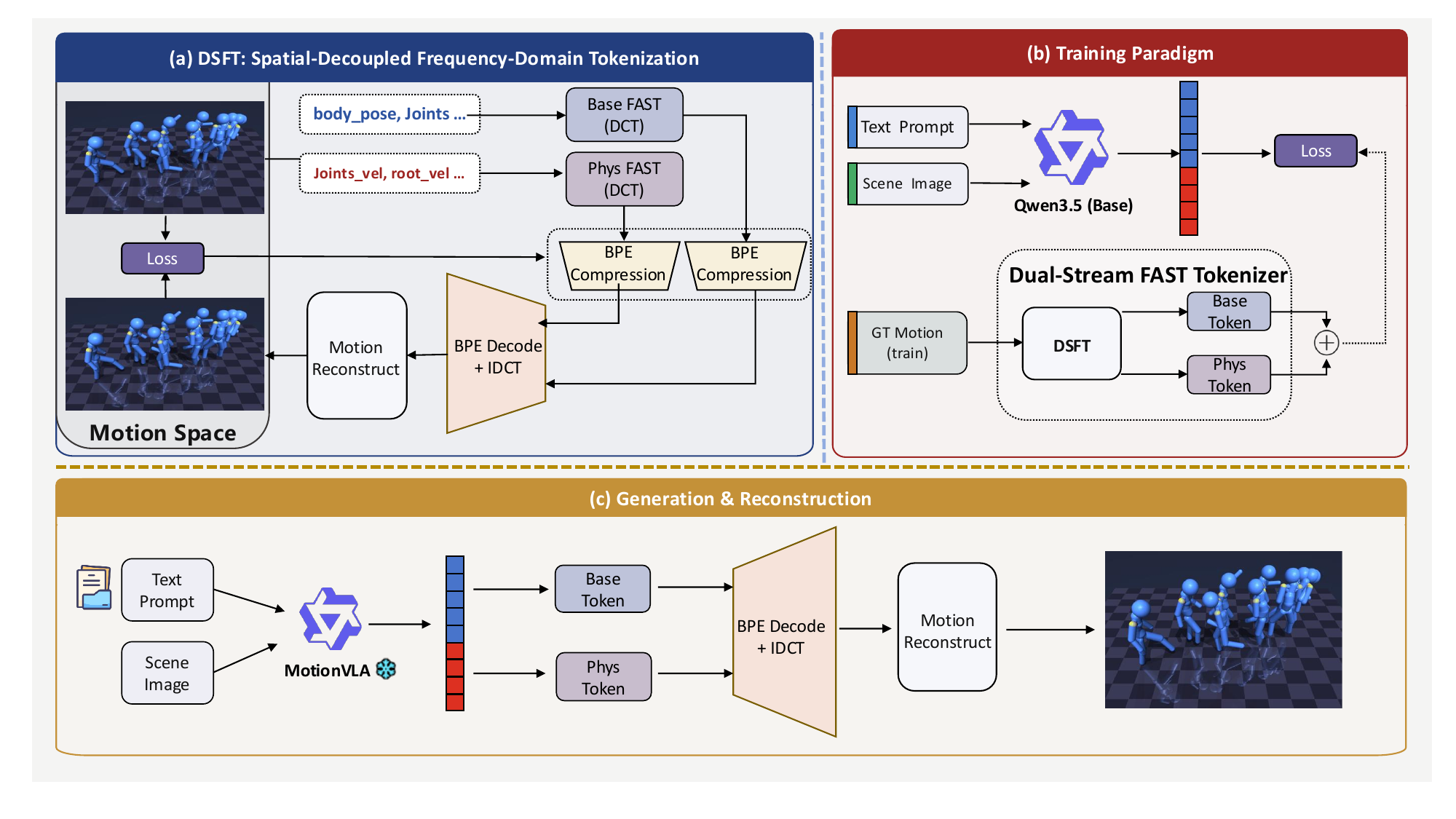}
  \caption{
Overview of MotionVLA. (a) DSFT performs dual-stream frequency tokenization by decomposing motion into Base and Phys components and converting them into discrete tokens. (b) During training, MotionVLA learns to autoregressively predict the unified motion token sequence under text and scene-image conditioning, supervised by DSFT tokens derived from ground-truth motion. (c) At inference time, the model generates Base and Phys tokens conditioned on multimodal inputs, which are then decoded and recombined to reconstruct the final motion sequence.
}
  \label{fig:overview}
  \vspace{-0.4cm}
\end{figure*}

As illustrated in Figure~\ref{fig:overview}, MotionVLA comprises two key components: DSFT, a dual-stream frequency tokenizer that converts motion sequences into discrete Base and Phys token streams, and a Qwen3.5-based autoregressive backbone that generates these tokens conditioned on scene images and text instructions. Given a scene image $\mathbf{I}$ and a text description $\mathbf{t}$, the model encodes the multimodal context and autoregressively predicts a unified motion token sequence
$[M_\text{BOS},\, b_1,\ldots,b_N,\, M_\text{SEP},\, p_1,\ldots,p_M,\, M_\text{EOS}]$,
where Base tokens $b_i$ capture low-frequency pose semantics and Phys tokens $p_j$ encode high-frequency physical dynamics. After generation, the two streams are independently decoded through BPE inversion and inverse DCT, and then recombined to reconstruct the full motion sequence. This design preserves a unified autoregressive formulation while explicitly disentangling semantic structure from physical dynamics.

\subsection{DSFT: Dual-Stream Frequency-Domain Tokenizer}
\label{sec:dsfast}

\begin{figure}[ht]
  \centering
  \includegraphics[width=\linewidth]{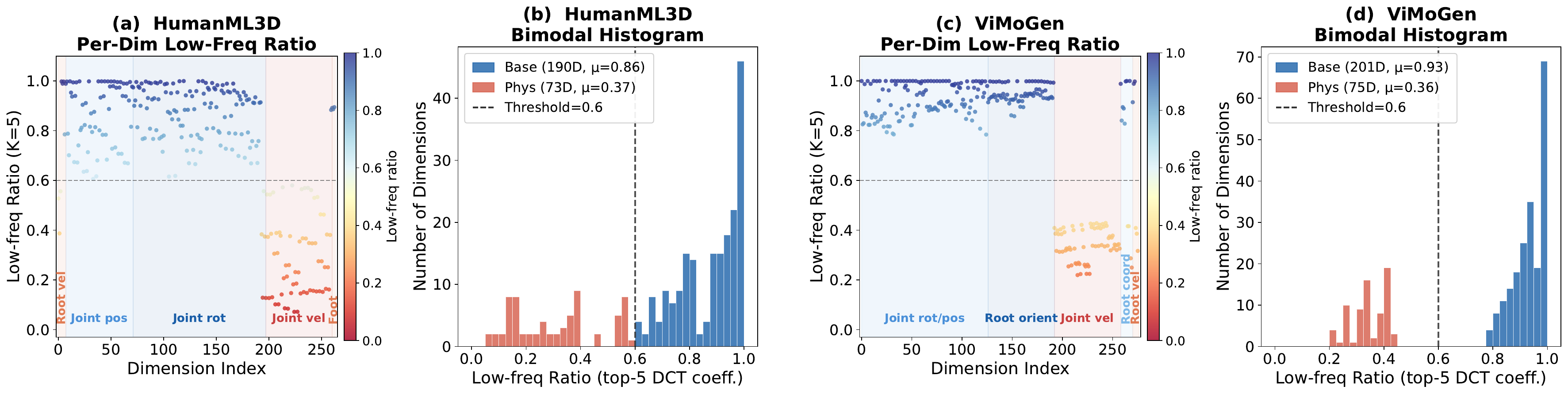}
  \caption{Frequency-domain clustering of motion dimensions.
(a) Per-dimension low-frequency ratio on HumanML3D.
(b) Corresponding histogram on HumanML3D.
(c/d) Corresponding plots on ViMoGen.
Both datasets exhibit a consistent bimodal separation between low-frequency Base dimensions and high-frequency Phys dimensions.}
  \label{fig:freq_clustering}
    \vspace{-0.4cm}
\end{figure}

DSFT is motivated by a simple observation: human motion is not spectrally homogeneous. Joint positions and rotations evolve relatively smoothly over time and are therefore dominated by low-frequency components, whereas joint velocities exhibit much stronger high-frequency behavior. This distinction follows naturally from the differentiation theorem, since temporal differentiation amplifies higher-frequency coefficients. To verify this property directly from data, we analyze the DCT energy distribution of each motion dimension on both HumanML3D (263 dimensions)  and ViMoGen (276 dimensions), and characterize each dimension by its low-frequency ratio, defined as the fraction of energy covered by the first five DCT coefficients. As shown in Figure~\ref{fig:freq_clustering}, the resulting distributions are strongly bimodal on both datasets, revealing a consistent separation between low-frequency semantic dimensions and high-frequency physical dimensions.

Based on this observation, we partition motion into two streams according to physical semantics. The Base stream contains position- and rotation-related dimensions that primarily encode pose semantics, whereas the Phys stream contains velocity-related dimensions that primarily encode physical dynamics. Concretely, this yields $(D_b,D_p)=(190,73)$ for HumanML3D and $(201,75)$ for ViMoGen
(see Appendix~\ref{app:dsfast_partition} for the exact per-field index mapping). We therefore represent a motion sequence $\mathbf{M}\in\mathbb{R}^{T\times D}$ as
\begin{equation}
  \mathbf{M}_{\text{base}} \in \mathbb{R}^{T \times D_b},
  \qquad
  \mathbf{M}_{\text{phys}} \in \mathbb{R}^{T \times D_p}.
\end{equation}

\begin{figure}[t]
  \centering
  \includegraphics[width=\linewidth]{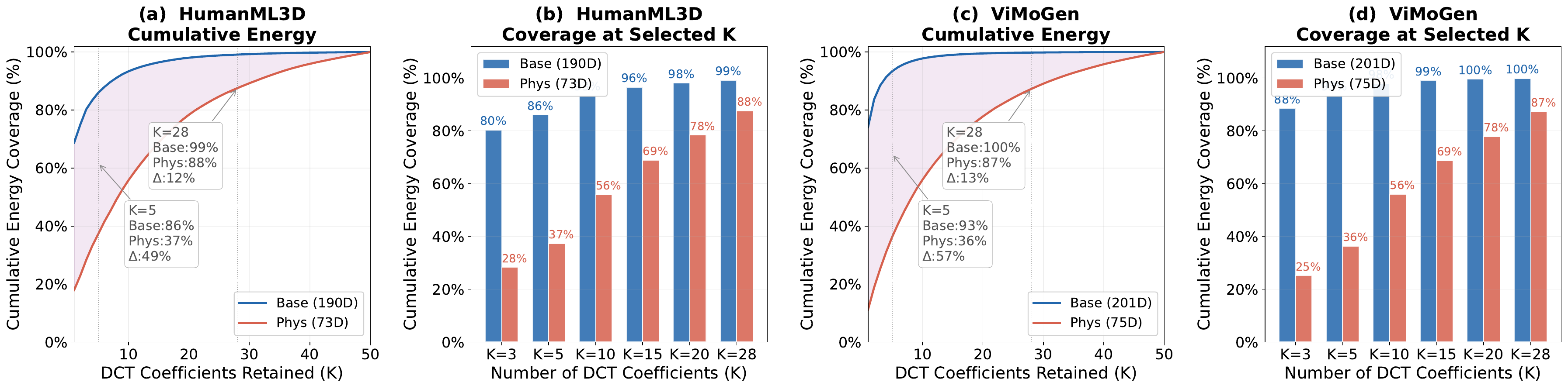}
  \caption{Energy coverage of the Base and Phys streams under different DCT truncation lengths.
(a,b) Results on HumanML3D.
(c,d) Corresponding results on ViMoGen.
The Base stream is highly compressible with small $K$, whereas the Phys stream requires substantially larger $K$ to preserve its energy.}
  \label{fig:energy_coverage}
  \vspace{-0.4cm}
\end{figure}

This distinction is further quantified in Figure~\ref{fig:energy_coverage}, which compares the cumulative DCT energy retained by the Base and Phys streams as the number of preserved coefficients increases. As shown in the figure, the Base stream is highly compressible: retaining only $K=5$ coefficients already covers about 86\% to 93\% of its energy across datasets. In contrast, the Phys stream is substantially more broadband, with the same $K=5$ covering only about 37\% of its energy. Therefore, compressing both streams under a shared frequency budget would inevitably favor the low-frequency Base stream while discarding a large portion of high-frequency physical information. This also explains why single-stream tokenization tends to preserve pose structure more easily than physical dynamics, leading to systematic information loss in the latter.

Accordingly, we retain different numbers of DCT coefficients for the two streams, using a small truncation length for the Base stream and a larger one for the Phys stream. Specifically, we set $K_b=5$ and $K_p=25$, and apply DCT independently to obtain
\begin{equation}
  \mathbf{C}_{\text{base}} = \operatorname{DCT}(\mathbf{M}_{\text{base}})_{[:K_b]},
  \qquad
  \mathbf{C}_{\text{phys}} = \operatorname{DCT}(\mathbf{M}_{\text{phys}})_{[:K_p]}.
\end{equation}
After truncation, each stream is flattened and encoded by an independently trained BPE tokenizer, yielding a Base token sequence $\mathbf{b}$ and a Phys token sequence $\mathbf{p}$. During decoding, we first recover the truncated coefficients by inverse BPE mapping, and then reconstruct the two time-domain streams by inverse DCT. Finally, the reconstructed Base and Phys streams are concatenated along the feature dimension to recover the complete motion sequence.

In this way, DSFT converts continuous motion into two complementary token streams, which provide the foundation for the unified autoregressive generation framework described next.

\subsection{MotionVLA: Unified Sequence Formulation, Objective, and Inference}
\label{sec:hier}

Given a scene image $\mathbf{I}$ and a text description $\mathbf{t}$, MotionVLA formulates motion generation as a unified autoregressive sequence modeling problem. Specifically, each motion sample is represented as
\begin{equation}
  \mathbf{s} =
  [M_\text{BOS},\, b_1,\ldots,b_N,\, M_\text{SEP},\, p_1,\ldots,p_M,\, M_\text{EOS}],
\end{equation}
where $b_i$ denotes Base tokens and $p_j$ denotes Phys tokens. In this way, MotionVLA preserves a simple autoregressive formulation while imposing a structured semantic-to-physical generation order.

This ordering is important because physical dynamics are typically conditioned on the underlying pose structure. By placing Phys tokens after all Base tokens, MotionVLA allows each Phys prediction to attend to the complete preceding Base context through causal attention. As a result, the model can generate physical dynamics with access to the full semantic pose information, rather than predicting both streams in an entangled manner. This design enables a hierarchical semantic-to-physical generation process within a standard autoregressive transformer.

To instantiate this sequence within the backbone token space, we extend the original vocabulary with motion tokens and three structural markers, yielding
\begin{equation}
V = V_{\text{LM}} + V_{\text{motion}} + 3,
\qquad
V_{\text{motion}} = V_{\text{base}} + V_{\text{phys}},
\end{equation}
where $V_{\text{LM}}$ denotes the original backbone vocabulary, and $V_{\text{motion}}$ denotes the motion vocabulary induced by DSFT.

During training, the scene image and text description serve as conditioning context, while the model is optimized to predict the motion portion of the sequence with teacher forcing. Let $\mathbf{y}$ denote the training targets, where non-motion positions are masked out. We optimize MotionVLA with a masked next-token prediction objective:
\begin{equation}
\mathcal{L}_{\text{train}}
=
\operatorname{CE}\!\left(\mathbf{z} + \mathbf{m},\, \mathbf{y}\right),
\end{equation}
where $\mathbf{z}$ denotes the output logits, and $\mathbf{m}$ is a logit mask that restricts prediction to valid motion tokens and structural markers. This objective prevents probability mass from being assigned to irrelevant vocabulary entries and focuses learning on the motion token space.

During inference, we further impose a phase-aware generation constraint to preserve the intended Base-to-Phys order. Before generating $M_\text{SEP}$, only Base tokens and $M_\text{SEP}$ are allowed; after $M_\text{SEP}$ is produced, only Phys tokens and $M_\text{EOS}$ are allowed. This phase-aware mask ensures that semantic structure is generated before physical dynamics, while preventing the model from mixing the two streams during decoding. The complete inference procedure is summarized in Algorithm~\ref{alg:inference}.

\begin{algorithm}[t]
\caption{Phase-Aware Autoregressive Generation in MotionVLA}
\label{alg:inference}
\begin{algorithmic}[1]
\Require Scene image $\mathbf{I}$, text instruction $\mathbf{t}$,
         trained MotionVLA, phase-aware logit mask $\mathbf{m}$
\Ensure  Reconstructed motion sequence $\mathbf{M} \in \mathbb{R}^{T \times D}$
\State Encode $\mathbf{I}$, $\mathbf{t}$ through Qwen3.5 $\rightarrow$ context representations
\State Initialize token buffer $\mathbf{s} \leftarrow [M_\text{BOS}]$,\; $\textit{phase} \leftarrow \textsc{Base}$
\While{last token $\neq M_\text{EOS}$}
    \If{$\textit{phase} = \textsc{Base}$}
        \State Apply mask: allow $\{b_i\} \cup \{M_\text{SEP}\}$ only
    \Else
        \State Apply mask: allow $\{p_j\} \cup \{M_\text{EOS}\}$ only
    \EndIf
    \State Sample next token $\tau \sim \operatorname{softmax}(\text{logits} + \mathbf{m})$
    \State Append $\tau$ to $\mathbf{s}$
    \If{$\tau = M_\text{SEP}$}
        \State $\textit{phase} \leftarrow \textsc{Phys}$
    \EndIf
\EndWhile
\State Extract Base tokens $\mathbf{b}$ and Phys tokens $\mathbf{p}$ from $\mathbf{s}$
\State $\mathbf{M}_\text{base} \leftarrow \operatorname{IDCT}\!\left(\operatorname{BPE}^{-1}(\mathbf{b})\right)$
\State $\mathbf{M}_\text{phys} \leftarrow \operatorname{IDCT}\!\left(\operatorname{BPE}^{-1}(\mathbf{p})\right)$
\State \Return $\mathbf{M} \leftarrow [\mathbf{M}_\text{base} \;\|\; \mathbf{M}_\text{phys}]$
  \Comment{concatenate along spatial dimension}
\end{algorithmic}
\end{algorithm}

\section{Experiments}
\label{sec:experiments}

\subsection{Datasets and Evaluation Metrics}

We conduct experiments on two settings.
In the first setting, we train on \textbf{ViMoGen-228K}~\cite{vimogen}, a large-scale
multimodal motion dataset, and evaluate on \textbf{MBench}~\cite{vimogen},
the associated fine-grained physical quality benchmark.
In the second setting, we train and evaluate on \textbf{HumanML3D}~\cite{humanml3d}
under the standard text-to-motion protocol.
Table~\ref{tab:datasets} summarizes dataset statistics;
detailed metric definitions and evaluation protocols are provided in
Appendix~\ref{sec:appendix_metrics}.

\begin{table}[ht]
\centering
\small
\setlength{\tabcolsep}{5pt}
\caption{\textbf{Dataset statistics.}
ViMoGen-228K is the training dataset; MBench is its associated evaluation benchmark (not a dataset subset).
$\dagger$: optical motion capture with marker-based GT.
$\ddagger$: in-the-wild video with pseudo-GT from SMPL estimation.
$\#$: generative synthesis with physics-based renderer.}
\label{tab:datasets}
\begin{tabular}{lrrcccc}
\toprule
\textbf{Dataset / Benchmark} & \textbf{\#Clip} & \textbf{\#Hour} & \textbf{Scene} & \textbf{Motion} & \textbf{Video} & \textbf{Role} \\
\midrule
HumanML3D~\cite{humanml3d}                       & 14,616  &  28.6 & Indoor       & GT        & --         & Train / Eval \\
\midrule
ViMoGen-228K~\cite{vimogen}                                     & 228,236 & 369.4 & Mixed        & Mixed     & --         & Train \\
\quad $\bullet$ Optical MoCap$^\dagger$           & 171,542 & 292.7 & Indoor       & GT        & --         & Train\\
\quad $\bullet$ In-the-Wild Video$^\ddagger$      &  41,971 &  61.4 & In-the-Wild  & Pseudo GT & \checkmark &  Train\\
\quad $\bullet$ Synthetic Video$^\#$              &  14,723 &  16.6 & In-the-Wild  & Pseudo GT & \checkmark & Not used \\
MBench~\cite{vimogen}                            & 450     & --    & In-the-Wild  & GT        & \checkmark & Eval only \\
\bottomrule
\end{tabular}
\vspace{-0.4cm}
\end{table}

\subsection{Baselines}

We compare MotionVLA with representative baselines spanning three paradigms:
discrete autoregressive generation, diffusion-based methods, and approaches
with improved motion tokenization, as summarized in Table~\ref{tab:baselines}.
On ViMoGen-228K and MBench, prior methods are evaluated under their original
text-driven setting, whereas MotionVLA additionally conditions on the scene image.
On HumanML3D, all methods follow the same standard text-to-motion protocol
for a fair comparison.

\subsection{Experimental Setup}
Prior to model training, we train the DSFT tokenizer independently on each benchmark's training split, ensuring that the discrete motion representation is adapted to the motion statistics of each dataset. Full training details and feature partition specifications are provided in Appendix~\ref{app:impl} and~\ref{app:dsfast_partition}.

We then conduct four groups of experiments to evaluate MotionVLA comprehensively: a main benchmark evaluation on MBench for scene-conditioned motion generation, a text-to-motion generalization evaluation on HumanML3D, a DSFT tokenizer reconstruction analysis to assess representation quality, and an ablation study to examine the impact of key design choices. Detailed hyperparameter settings are provided in Appendix~\ref{app:impl}.

\subsection{Main Results on MBench}
\label{sec:main_results}

Table~\ref{tab:main_mbench} reports the quantitative comparison on MBench,
which evaluates models trained on ViMoGen-228K across eight fine-grained quality
dimensions.
Despite using a lightweight 2B backbone, MotionVLA achieves the best results on
Motion-Condition Consistency and Foot Sliding, while ranking second on
Motion Generalizability and Jitter Degree.
These gains indicate that the proposed framework is particularly effective at
improving multimodal condition alignment and suppressing local temporal artifacts.

At inference time, the target motion length T is provided externally, and MotionVLA generates DSFT tokens conditioned on this target horizon. This pattern is consistent with our design.
Scene-aware conditioning mainly improves semantic alignment, while the DSFT
dual-stream tokenizer reduces local temporal artifacts such as jitter and foot sliding.
Meanwhile, MotionVLA does not dominate all physical metrics, indicating that
low-level geometric quality remains challenging under the smaller model scale.
Overall, MotionVLA provides a favorable trade-off between multimodal controllability
and physical motion quality on MBench.

\begin{table}[t]
\centering
\small
\setlength{\tabcolsep}{3pt}
\caption{\textbf{MotionVLA~\cite{vimogen} evaluation on MBench.}
         $\uparrow$: higher is better; $\downarrow$: lower is better.
         $\dagger$: uses additional visual (scene) input.
         Best in \textbf{bold}, second best \underline{underlined}.}
\label{tab:main_mbench}
\resizebox{\columnwidth}{!}{%
\begin{tabular}{lcccccccc}
\toprule
\textbf{Method}
  & \textbf{\shortstack{Motion-Cond.\\Consistency}} $\uparrow$
  & \textbf{\shortstack{Motion\\Generaliz.}} $\uparrow$
  & \textbf{\shortstack{Jitter\\Degree}} $\downarrow$
  & \textbf{\shortstack{Dynamic\\Degree}} $\uparrow$
  & \textbf{\shortstack{Foot\\Floating}} $\downarrow$
  & \textbf{\shortstack{Foot\\Sliding}} $\downarrow$
  & \textbf{\shortstack{Body\\Penetration}} $\downarrow$
  & \textbf{\shortstack{Pose\\Quality}} $\downarrow$ \\
\midrule
MDM~\cite{mdm} {\scriptsize(ICLR'23)}
  & 0.42 & 0.51 & 0.0136 & 0.0376 & 0.156 & 0.0136 & 1.68 & 2.67 \\
T2M-GPT~\cite{t2m_gpt} {\scriptsize(CVPR'23)}
  & 0.39 & 0.38 & 0.0156 & 0.0349 & 0.209 & 0.0156 & 1.33 & 2.43 \\
FineMoGen~\cite{finemogen} {\scriptsize(NeurIPS'24)}
  & 0.37 & 0.42 & 0.0118 & 0.0386 & 0.281 & 0.0091 & \underline{1.18} & 2.28 \\
MotionLCM~\cite{motionlcm} {\scriptsize(ECCV'24)}
  & 0.48 & 0.55 & 0.0218 & \textbf{0.0439} & 0.193 & 0.0202 & 1.73 & 2.40 \\
MoMask~\cite{momask} {\scriptsize(CVPR'24)}
  & 0.38 & 0.44 & 0.0147 & 0.0396 & 0.178 & 0.0147 & 1.48 & 2.67 \\
MotionDiffuse~\cite{motiondiffuse} {\scriptsize(TPAMI'24)}
  & 0.44 & 0.42 & 0.0111 & 0.0289 & \textbf{0.126} & 0.0063 & 1.35 & 2.21 \\
MotionCraft~\cite{motioncraft} {\scriptsize(CVPR'25)}
  & 0.42 & 0.45 & 0.0132 & \underline{0.0420} & 0.402 & 0.0090 & \textbf{1.15} & \underline{2.12} \\
ViMoGen~\cite{vimogen} {\scriptsize(ICLR'26)}
  & \underline{0.53} & \textbf{0.68} & \textbf{0.0108} & 0.0251 & 0.204 & 0.0064 & 1.78 & 2.38 \\
ViMoGen-light~\cite{vimogen} {\scriptsize(ICLR'26)}
  & 0.47 & 0.55 & 0.0129 & 0.0294 & 0.155 & \underline{0.0051} & 1.43 & \textbf{2.10} \\
\midrule
\textbf{MotionVLA (Ours)}$^\dagger$
  & \textbf{0.55} & \underline{0.66} & \underline{0.0110} & 0.0419
  & \underline{0.149} & \textbf{0.0049} & 1.34 & 2.14 \\
\bottomrule
\end{tabular}%
}
\vspace{-0.4cm}
\end{table}

\subsection{Results on HumanML3D}
\label{sec:humanml3d_results}
Table~\ref{tab:main_humanml3d} reports text-to-motion generation results on
HumanML3D under the standard benchmark setting.
Although MotionVLA uses a lightweight 2B backbone and is designed for multimodal
motion generation, it remains competitive on this purely text-driven benchmark,
achieving the Diversity score closest to the real data distribution and the
highest MModality among generated methods.
Its R-Precision, FID, and MM-Dist scores also remain competitive with strong
recent baselines, indicating that the proposed framework transfers beyond the
multimodal training setting.

This pattern is consistent with our design.
Since HumanML3D removes visual conditioning, the gains mainly reflect the motion
representation itself rather than scene input.
By separating low-frequency motion semantics from high-frequency physical dynamics,
DSFT preserves richer motion variation while maintaining competitive sample
fidelity and text-motion alignment.
Overall, these results show that MotionVLA generalizes effectively beyond ViMoGen
and provides a strong diversity-quality trade-off even at a relatively small
2B model scale.
\begin{table}[t]
\centering
\small
\setlength{\tabcolsep}{4pt}
\caption{\textbf{Text-to-motion results on HumanML3D.}
$\uparrow$: higher is better; $\downarrow$: lower is better; $\rightarrow$: closer to real is better.
For Diversity, best and second best are determined by the distance to the Real score.
$\ddagger$: GenM3 uses a retrained evaluator on 30\,FPS data; GenM3$^*$ uses only HumanML3D text pairs.
$\S$: DisCoRD is applied on top of MoMask; Diversity is not reported in the original paper.}
\label{tab:main_humanml3d}
\resizebox{\columnwidth}{!}{%
\begin{tabular}{lccccccc}
\toprule
\textbf{Method}
  & \textbf{R-P Top-1} $\uparrow$
  & \textbf{R-P Top-2} $\uparrow$
  & \textbf{R-P Top-3} $\uparrow$
  & \textbf{FID} $\downarrow$
  & \textbf{MM-Dist} $\downarrow$
  & \textbf{Diversity} $\rightarrow$
  & \textbf{MModality} $\uparrow$ \\
\midrule
\quad Real
  & $0.511^{{\pm}.003}$ & $0.703^{{\pm}.003}$ & $0.797^{{\pm}.002}$
  & $0.002^{{\pm}.000}$ & $2.974^{{\pm}.008}$ & $9.503^{{\pm}.065}$ & -- \\
\midrule
\multicolumn{8}{l}{\textit{Motion generation only}} \\
\quad TEMOS~\cite{temos} {\scriptsize(ECCV'22)}
  & $0.424^{{\pm}.002}$ & $0.612^{{\pm}.002}$ & $0.722^{{\pm}.002}$
  & $3.734^{{\pm}.028}$ & $3.703^{{\pm}.008}$ & $8.973^{{\pm}.071}$ & $0.368^{{\pm}.018}$ \\
\quad TM2T~\cite{tm2t} {\scriptsize(ECCV'22)}
  & $0.424^{{\pm}.003}$ & $0.618^{{\pm}.003}$ & $0.729^{{\pm}.002}$
  & $1.501^{{\pm}.017}$ & $3.467^{{\pm}.011}$ & $8.589^{{\pm}.076}$ & $2.424^{{\pm}.093}$ \\
\quad Guo et al.~\cite{humanml3d} {\scriptsize(CVPR'22)}
  & $0.455^{{\pm}.003}$ & $0.636^{{\pm}.003}$ & $0.736^{{\pm}.002}$
  & $1.087^{{\pm}.021}$ & $3.347^{{\pm}.008}$ & $9.175^{{\pm}.083}$ & $2.219^{{\pm}.074}$ \\
\quad MDM~\cite{mdm} {\scriptsize(ICLR'23)}
  & -- & -- & $0.611^{{\pm}.007}$
  & $0.544^{{\pm}.044}$ & $5.566^{{\pm}.027}$ & -- & \underline{$2.799^{{\pm}.072}$} \\
\quad MotionDiffuse~\cite{motiondiffuse} {\scriptsize(TPAMI'24)}
  & $0.491^{{\pm}.001}$ & $0.681^{{\pm}.001}$ & $0.782^{{\pm}.001}$
  & $0.630^{{\pm}.001}$ & $3.113^{{\pm}.001}$ & \underline{$9.410^{{\pm}.049}$} & $1.553^{{\pm}.042}$ \\
\quad T2M-GPT~\cite{t2m_gpt} {\scriptsize(CVPR'23)}
  & $0.492^{{\pm}.003}$ & $0.679^{{\pm}.002}$ & $0.775^{{\pm}.002}$
  & $0.141^{{\pm}.004}$ & $3.121^{{\pm}.009}$ & $9.722^{{\pm}.082}$ & $1.831^{{\pm}.048}$ \\
\quad FineMoGen~\cite{finemogen} {\scriptsize(NeurIPS'24)}
  & $0.504^{{\pm}.002}$ & $0.690^{{\pm}.002}$ & $0.784^{{\pm}.002}$
  & $0.151^{{\pm}.008}$ & $2.998^{{\pm}.008}$ & $9.263^{{\pm}.094}$ & $2.696^{{\pm}.079}$ \\
\quad MoMask~\cite{momask} {\scriptsize(CVPR'24)}
  & \underline{$0.521^{{\pm}.002}$} & \underline{$0.713^{{\pm}.002}$} & \underline{$0.807^{{\pm}.002}$}
  & \underline{$0.045^{{\pm}.002}$} & $2.958^{{\pm}.008}$ & -- & $1.241^{{\pm}.040}$ \\
\quad MotionGPT~\cite{motiongpt_aaai} {\scriptsize(AAAI'24)}
  & -- & -- & --
  & $0.567$ & $3.775$ & $9.006$ & -- \\
\quad DisCoRD~\cite{discord} {\scriptsize(ICCV'25)}$^\S$
  & \textcolor{red}{$0.524^{{\pm}.003}$} & \textcolor{red}{$0.715^{{\pm}.003}$} & \textcolor{red}{$0.809^{{\pm}.002}$}
  & \textcolor{red}{$0.032^{{\pm}.002}$} & $2.938^{{\pm}.010}$ & -- & $1.288^{{\pm}.043}$ \\
\quad GenM3$^*$~\cite{genm3} {\scriptsize(ICCV'25)}$^\ddagger$
  & $0.510^{{\pm}.002}$ & $0.702^{{\pm}.002}$ & $0.802^{{\pm}.002}$
  & $0.053^{{\pm}.002}$ & \underline{$2.860^{{\pm}.009}$} & $9.629^{{\pm}.077}$ & -- \\
\quad GenM3~\cite{genm3} {\scriptsize(ICCV'25)}$^\ddagger$
  & $0.511^{{\pm}.003}$ & $0.705^{{\pm}.002}$ & $0.804^{{\pm}.002}$
  & $0.046^{{\pm}.002}$ & \textcolor{red}{$2.852^{{\pm}.009}$} & $9.675^{{\pm}.087}$ & -- \\
\quad MG-MotionLLM~\cite{mg_motionllm} {\scriptsize(CVPR'25)}
  & $0.516^{{\pm}.002}$ & $0.706^{{\pm}.002}$ & $0.802^{{\pm}.003}$
  & $0.303^{{\pm}.010}$ & $2.952^{{\pm}.009}$ & $9.960^{{\pm}.073}$ & $2.125^{{\pm}.159}$ \\
\midrule
\textbf{MotionVLA (Ours)}
  & $0.507^{{\pm}.002}$ & $0.699^{{\pm}.002}$ & $0.798^{{\pm}.002}$
  & $0.071^{{\pm}.003}$ & $2.906^{{\pm}.009}$ & \textcolor{red}{$9.548^{{\pm}.081}$} & \textcolor{red}{$2.821^{{\pm}.091}$} \\
\bottomrule
\end{tabular}%
}
\vspace{-0.4cm}
\end{table}

\subsection{DSFT Tokenizer Analysis}
\label{sec:tokenizer_analysis}
We analyze DSFT on HumanML3D through controlled comparisons within the DCT+BPE
family.
Compared with a single-stream DCT+BPE baseline, DSFT produces a more compact
token sequence and a substantially lower reconstruction Fréchet inception distance
(rFID), despite having higher reconstruction root mean square error (rRMSE) and MPJPE.
This suggests that lower pointwise reconstruction error does not necessarily imply
better tokenizer quality.

As the Phys-stream truncation length $K_p$ increases, rRMSE, MPJPE, and rFID all
improve consistently.
We use $K_p{=}25$ as the default setting because it already offers a strong
compactness-fidelity trade-off while matching the main tokenizer configuration.

\begin{table}[t]
\centering
\small
\caption{\textbf{DSFT tokenizer reconstruction analysis on HumanML3D.}
         Smaller Tok./Frame indicates a more compact tokenization;
         lower rRMSE, MPJPE, and rFID are better.}
\label{tab:tokenizer_analysis}
\begin{tabular}{lcccc}
\toprule
\textbf{Method} & \textbf{Tok./Frame} & \textbf{rRMSE} & \textbf{MPJPE} & \textbf{rFID} \\
\midrule
Single-Stream DCT+BPE & 15.21 & 0.0164 & 0.0054 & 0.9461 \\
DSFT ($K_p{=}25$)  & 11.24 & 0.0226 & 0.0093 & 0.1868 \\
\midrule
DSFT ($K_p{=}10$)  &  8.80 & 0.0240 & 0.0108 & 1.3404 \\
DSFT ($K_p{=}15$)  &  9.74 & 0.0233 & 0.0102 & 0.5966 \\
DSFT ($K_p{=}20$)  & 10.55 & 0.0228 & 0.0097 & 0.2895 \\
DSFT ($K_p{=}25$)  & 11.24 & 0.0226 & 0.0093 & 0.1868 \\
DSFT ($K_p{=}30$)  & 11.81 & 0.0224 & 0.0091 & 0.1380 \\
\bottomrule
\end{tabular}
\vspace{-0.4cm}
\end{table}

Table~\ref{tab:tokenizer_analysis} also shows that increasing the Phys-stream
truncation length $K_p$ consistently improves reconstruction quality.
As $K_p$ increases from 10 to 30, both rRMSE and MPJPE decrease, while rFID drops
from 1.340 to 0.138.
We therefore use $K_p{=}25$ as the default setting, since it already provides a
strong compactness-fidelity trade-off while matching the main tokenizer
configuration.

\subsection{Ablation Studies}
\label{sec:ablation}

We conduct two ablation studies on MBench to examine the effect of backbone scale
and the Phys-stream truncation length $K_p$.
Since HumanML3D does not provide visual observations, we focus the ablation
analysis on the scene-conditioned ViMoGen-228K--MBench setting.

\noindent\textbf{Backbone Scale.}
Table~\ref{tab:ablation_scale} shows that scaling up the Qwen3.5 backbone yields consistent but diminishing gains on MBench. The largest improvement comes from 0.8B to 2B, while the gains from 2B to 4B and 9B are relatively small. For some metrics, the 2B and 4B models appear unchanged after rounding, since only two decimal places are reported in those columns. One possible explanation is that, under the current data scale and training recipe, the available supervision is not sufficient to fully exploit substantially larger backbones. Moreover, because MotionVLA predicts a fixed DSFT tokenization rather than continuous motion directly, the effective information carried by the token representation may also limit how much additional capacity can be translated into measurable gains. As a result, a 2B model already captures most of the achievable improvement in the current setting, making it a favorable default choice in terms of both performance and efficiency.
\begin{table}[t]
\centering
\small
\setlength{\tabcolsep}{4pt}
\caption{\textbf{Backbone scale ablation on MBench.}
         $\dagger$: default configuration used in main experiments.
         $\uparrow$/$\downarrow$: higher/lower is better.}
\label{tab:ablation_scale}
\resizebox{\columnwidth}{!}{%
\begin{tabular}{lc cccccccc}
\toprule
\textbf{Backbone}
  & \textbf{Params}
  & \textbf{\shortstack{Motion-Cond.\\Consistency}} $\uparrow$
  & \textbf{\shortstack{Motion\\Generaliz.}} $\uparrow$
  & \textbf{\shortstack{Jitter\\Degree}} $\downarrow$
  & \textbf{\shortstack{Dynamic\\Degree}} $\uparrow$
  & \textbf{\shortstack{Foot\\Floating}} $\downarrow$
  & \textbf{\shortstack{Foot\\Sliding}} $\downarrow$
  & \textbf{\shortstack{Body\\Penetration}} $\downarrow$
  & \textbf{\shortstack{Pose\\Quality}} $\downarrow$ \\
\midrule
Qwen3.5-0.8B                  & 0.8B & 0.51 & 0.60 & 0.0122 & 0.0364 & 0.162 & 0.0058 & 1.46 & 2.26 \\
\textbf{Qwen3.5-2B}$^\dagger$ & \textbf{2B} & 0.55 & 0.66 & 0.0110 & 0.0419 & 0.149 & 0.0049 & 1.34 & 2.14 \\
Qwen3.5-4B                    & 4B   & 0.55 & 0.66 & 0.0109 & 0.0393 & 0.146 & 0.0049 & 1.31 & 2.12 \\
Qwen3.5-9B                    & 9B   & 0.56 & 0.68 & 0.0107 & 0.0396 & 0.144 & 0.0047 & 1.30 & 2.09 \\
\bottomrule
\end{tabular}}
\vspace{-0.4cm}
\end{table}

\noindent\textbf{DSFT Truncation Parameter $K_p$.}
Table~\ref{tab:ablation_kp} studies the effect of the Phys-stream truncation
length $K_p$ while fixing $K_b{=}5$ under the default 2B backbone.
As $K_p$ increases, Phys-stream energy coverage improves consistently, indicating
that a larger frequency budget preserves more high-frequency physical dynamics.
From $K_p{=}10$ to $K_p{=}25$, this added physical capacity is accompanied by
clear improvements on most MBench metrics, showing that richer physical signals
benefit overall motion quality.
However, a larger $K_p$ also increases the motion sequence length, and the gains
do not continue monotonically at $K_p{=}30$, where several metrics slightly
degrade.
We therefore use $K_p{=}25$ as the default setting, as it provides the best
overall balance between physical detail preservation and sequence efficiency
under the current 2B model scale.

\begin{table}[t]
\centering
\small
\setlength{\tabcolsep}{4pt}
\caption{\textbf{DSFT $K_p$ ablation on MBench.}
         $K_b{=}5$ fixed. Tok./Sample: average motion token count.
         $\uparrow$/$\downarrow$: higher/lower is better.}
\label{tab:ablation_kp}
\resizebox{\columnwidth}{!}{%
\begin{tabular}{ccc cccccccc}
\toprule
$K_p$ & \textbf{\shortstack{Phys\\Coverage}} $\uparrow$ & \textbf{Tok./Sample}
  & \textbf{\shortstack{Motion-Cond.\\Consistency}} $\uparrow$
  & \textbf{\shortstack{Motion\\Generaliz.}} $\uparrow$
  & \textbf{\shortstack{Jitter\\Degree}} $\downarrow$
  & \textbf{\shortstack{Dynamic\\Degree}} $\uparrow$
  & \textbf{\shortstack{Foot\\Floating}} $\downarrow$
  & \textbf{\shortstack{Foot\\Sliding}} $\downarrow$
  & \textbf{\shortstack{Body\\Penetration}} $\downarrow$
  & \textbf{\shortstack{Pose\\Quality}} $\downarrow$ \\
\midrule
10          & $\sim$50\% & 438 & 0.49 & 0.58 & 0.0131 & 0.0338 & 0.171 & 0.0062 & 1.48 & 2.28 \\
15          & $\sim$63\% & 472 & 0.52 & 0.61 & 0.0121 & 0.0364 & 0.160 & 0.0054 & 1.41 & 2.21 \\
20          & $\sim$73\% & 507 & 0.54 & 0.63 & 0.0116 & 0.0375 & 0.154 & 0.0051 & 1.37 & 2.17 \\
\textbf{25} & $\sim$80\% & 541 & 0.55 & 0.66 & 0.0110 & 0.0419 & 0.149 & 0.0049 & 1.34 & 2.14 \\
30          & $\sim$85\% & 569 & 0.53 & 0.64 & 0.0113 & 0.0379 & 0.152 & 0.0052 & 1.38 & 2.16 \\
\bottomrule
\end{tabular}}
\vspace{-0.4cm}
\end{table}

\subsection{Simulation, Deployment and Human Preference Study}
\label{sec:sim_real}

To complement automatic metrics, we evaluate MotionVLA in MuJoCo simulation, deploy it on a Unitree G1 EDU humanoid under the text-to-motion setting, and conduct a blinded human preference study against ViMoGen~\cite{vimogen}. Given a text prompt, MotionVLA generates motion tokens that DSFT decodes into real-time joint-angle trajectories. Five domain experts assess 100 anonymized text-conditioned motion pairs, producing 500 pairwise comparisons. Detailed simulation, deployment, and evaluation protocols are provided in Appendix~\ref{app:simulation} and~\ref{sec:human_study}.

As summarized in Table~\ref{tab:user_study_2}, MotionVLA is preferred in 64.0\% of comparisons, compared with 14.0\% for ViMoGen and 22.0\% ties, indicating a clear perceptual advantage in overall motion quality.
\begin{table}[t]
\centering
\small
\setlength{\tabcolsep}{5pt}
\caption{\textbf{Human preference study} (\%) on 100 prompts $\times$ 5 experts.
         Ours: MotionVLA preferred; Tie: comparable; Base: ViMoGen preferred.}
\label{tab:user_study_2}
\begin{tabular}{lcccccc}
\toprule
 & \textbf{Expert 1} & \textbf{Expert 2} & \textbf{Expert 3} & \textbf{Expert 4} & \textbf{Expert 5} & \textbf{Avg.} \\
\midrule
\textbf{Ours (MotionVLA preferred)}  & 72 & 58 & 69 & 55 & 66 & \textbf{64.0} \\
\textbf{Tie (Same taste)}   & 17 & 26 & 19 & 28 & 20 & 22.0 \\
\textbf{Base (Vimogen preferred)}  & 11 & 16 & 12 & 17 & 14 & 14.0 \\
\bottomrule
\end{tabular}
\vspace{-0.4cm}
\end{table}

\section{Discussions and Conclusions}

In this work, we address humanoid motion generation through coordinated innovations in tokenizer design, autoregressive modeling, and evaluation. (1) We introduce DSFT, a dual-stream frequency-domain tokenizer that decomposes motion into Base and Phys streams, motivated by the observation that pose-related and dynamic signals exhibit different spectral characteristics and therefore should not be forced into a single shared tokenization space. By assigning separate frequency budgets to the two streams, DSFT better preserves both motion semantics and high-frequency physical dynamics. (2) Built on top of this tokenizer, MotionVLA adapts a standard vision-language autoregressive backbone to unified motion generation, showing that multimodal controllability and physical motion quality can be improved within a simple sequence modeling framework. (3) Experiments on HumanML3D and ViMoGen--MBench show strong performance on both automatic metrics and human preference evaluation, supporting the effectiveness of frequency-aware dual-stream tokenization for both multimodal generation and transfer to standard text-to-motion settings. More broadly, these results suggest that effective motion tokenization depends not only on compression or pointwise reconstruction quality, but also on how motion signals are organized before discretization. A detailed discussion of related work is provided in Appendix~\ref{app:related}.

\noindent\textbf{Limitations and Future Work.}
Our current study focuses on a lightweight 2B backbone and a limited set of
benchmarks, and therefore does not yet support broader conclusions about scaling
behavior or cross-dataset generalization.
In addition, the current framework uses a fixed stream partition, fixed
truncation lengths, and a predefined Base-to-Phys generation order, which may
not be optimal for all motion types or sequence lengths.
Future work will extend the evaluation to larger backbones, broader datasets,
and more adaptive tokenization and dependency modeling schemes.

\bibliographystyle{plain}   %
\bibliography{bibliography}    %

\clearpage
\appendix

\section{Related Work}
\label{app:related}
\subsection{Human Motion Generation}

Text-driven human motion generation aims to synthesize realistic 3D human motion sequences from natural language descriptions. Mainstream methods build generation frameworks upon VQ-VAE discretization and autoregressive Transformers. T2M-GPT~\cite{t2m_gpt} first combines VQ-VAE with GPT next-token prediction, establishing the representative paradigm in this line of work. MotionGPT~\cite{motiongpt} treats motion as a foreign language and jointly trains multiple motion tasks under a unified language model. MoMask~\cite{momask} introduces Residual VQ (RVQ) hierarchical codebooks and Masked Transformers, reducing the FID to 0.045, while MG-MotionLLM~\cite{mg_motionllm} builds a multi-granularity motion-language framework with T5 as the backbone, extending semantic granularity from the sequence level to the segment level. GenM3~\cite{genm3} collects 11 datasets and employs a multi-expert VQ-VAE along with a multi-path Transformer to handle data heterogeneity, achieving a state-of-the-art FID of 0.035. MoSa~\cite{mosa} proposes RQ-VAE and a scalable autoregressive framework, outperforming the 10-step inference speed of MoMask by 27\%. More recently, MotionGPT3~\cite{motiongpt3} and LLaMo~\cite{llamo} shift towards continuous latent space autoregression to reduce the motion jitter caused by discrete quantization. Another line of work adopts diffusion models as the backbone. MDM~\cite{mdm} establishes the Transformer-based diffusion framework, and MotionDiffuse~\cite{motiondiffuse} extends it to support fine-grained control at the body-part level. FineMoGen~\cite{finemogen} achieves fine-grained spatio-temporal synthesis using Spatial-Temporal Mixed Attention (SAMI) and sparse MoE, while MotionStreamer~\cite{motionstreamer} combines diffusion with autoregression in a causal latent space for streaming generation. Recent systematic comparisons~\cite{rethink_diffusion} indicate that VQ-based autoregressive methods still hold an overall advantage on standard metrics such as FID and R-Precision.

\subsection{Motion Tokenization and Representation}

High-quality discrete motion representations form the basis of autoregressive generation methods. VQ-VAE~\cite{vqvae} introduces discrete bottlenecks into sequence modeling, enabling efficient compression and generation of continuous data. T2M-GPT~\cite{t2m_gpt} adapts this paradigm to the human motion domain, showing that motion can be effectively tokenized for text-driven generation. MoMask~\cite{momask} enhances hierarchical representation capabilities with Residual Vector Quantization (RVQ)~\cite{vqvae}, enabling multi-scale semantic abstraction.
In the frequency domain, FAST~\cite{fast} constructs an efficient robot action tokenizer by combining Discrete Cosine Transform (DCT) with Byte-Pair Encoding (BPE), showing that frequency-domain representations capture high-frequency fine-grained control signals more effectively. Following this direction, WaMo~\cite{wamo} applies wavelet multi-scale decomposition~\cite{wavelets} to motion trajectory analysis, further validating that multi-frequency decomposition improves fine-grained motion-text correspondence. Recent work also explores the integration of semantic guidance into the tokenization process: LG-Tok~\cite{lg_tok} proposes a language-guided Transformer tokenizer that introduces semantic alignment during the encoding stage, while
MoTok~\cite{motok} uses a diffusion decoder to decouple semantic abstraction from fine-grained reconstruction, maintaining high-fidelity reconstruction quality with single-layer tokens. DisCoRD~\cite{discord} approaches the problem from the decoding end, mapping discrete tokens back to continuous motion via rectified flow~\cite{rectified_flow} to partially reduce inter-frame jitter, yet leaves the structural cause (single-codebook quantization) intact at the representation level.
These methods collectively point to a core open problem: simultaneously capturing semantic structure (e.g., action labels, phase transitions) and physical dynamics (e.g., velocity, acceleration, contact forces) during tokenization. Since these two types of signals occupy overlapping frequency bands, disentangling them within a single quantization space is difficult. Prior works~\cite{momask,motok,discord} seek this balance within a unified representation; in contrast, DS-FAST resolves it orthogonally by separating the two signal types before quantization, thereby preserving both high-level structure and low-level motion quality.

\subsection{Vision-Language-Action Models}

Vision-Language-Action (VLA) models unify visual perception, language understanding, and action generation within a single end-to-end framework, representing a core research direction in embodied AI~\cite{vla_survey,embodied_ai}. Among representative works, OpenVLA~\cite{openvla}, trained with 7B parameters on approximately 970,000 multi-robot trajectories from the Open X-Embodiment dataset~\cite{openx}, demonstrates strong generalization across diverse robotic platforms and manipulation tasks. $\pi_0$~\cite{pi0} achieves end-to-end generation of 50 Hz high-frequency dexterous manipulation using a flow matching paradigm~\cite{flow_matching}, setting a new standard for fine manipulation tasks that demand precise temporal control.
Other notable efforts include RT-2~\cite{rt2}, which uses vision-language models (VLMs) for grounded robot control, and Octo~\cite{octo}, a generalist robot policy trained on large-scale multi-embodiment data. A recent survey~\cite{vla_survey} identifies that the quality of discrete action representations remains one of the core bottlenecks constraining the fine-grained control capabilities of VLAs, particularly in tasks requiring high-frequency feedback and multi-modal conditioning.
Unlike the aforementioned VLA works, which primarily focus on low-level robot control with short-horizon actions (typically <5 seconds), MotionVLA in this paper targets vision- and text-conditioned fine-grained human motion generation, a domain characterized by higher semantic complexity, longer temporal durations, and richer multimodal conditioning. Specifically, human motions are more diverse and nuanced than robotic actions, spanning a richer space of activities, emotions, and styles; motion sequences often span 10--30 seconds, requiring long-range temporal coherence; and generation must be grounded simultaneously in both visual context (e.g., scene layout, object affordances) and linguistic descriptions. By addressing these challenges, MotionVLA extends the VLA framework to the domain of human motion synthesis, connecting robotic action generation and human-centric animation.

\section{Detailed datasets and metrics}
\label{sec:appendix_metrics}

\paragraph{ViMoGen-228K and MBench.}
ViMoGen-228K~\cite{vimogen} is a large-scale multimodal motion dataset containing 228K motion sequences collected from three sources: optical motion capture, in-the-wild video annotation, and synthetic generation. In our experiments, MotionVLA is trained on the ViMoGen-228K training split and evaluated on MBench~\cite{vimogen}, following the official protocol. MBench contains 450 held-out prompts and reports eight fine-grained evaluation dimensions, including Motion-Condition Consistency, Motion Generalizability, Jitter Degree, Dynamic Degree, Foot Floating, Foot Sliding, Body Penetration, and Pose Quality.

\paragraph{HumanML3D.}
HumanML3D~\cite{humanml3d} is a standard benchmark for text-driven human motion generation. In our experiments, we train and evaluate MotionVLA on the official HumanML3D split under the standard text-to-motion setting. Because HumanML3D does not provide visual inputs, the model is used in text-only mode. Following prior work~\cite{humanml3d}, we report FID, R-Precision (Top-1/2/3), MM-Dist, Diversity, and MModality using the official pretrained feature extractor.

\begin{table}[ht]
\centering
\small
\setlength{\tabcolsep}{5pt}
\caption{\textbf{Baselines across three paradigms.}
         $\dagger$ denotes visual conditioning (MotionVLA only).
         All prior methods are text-driven.}
\label{tab:baselines}
\begin{tabular}{llll}
\toprule
\textbf{Method} & \textbf{Venue} & \textbf{Paradigm} & \textbf{Tokenizer} \\
\midrule
\multicolumn{4}{l}{\textit{Discrete Autoregressive}} \\
\quad T2M-GPT~\cite{t2m_gpt}           & CVPR 2023 & AR (GPT)       & VQ-VAE \\
\quad MoMask~\cite{momask}              & CVPR 2024 & Masked AR      & RVQ \\
\quad MG-MotionLLM~\cite{mg_motionllm} & CVPR 2025 & AR (T5)        & VQ \\
\quad GenM3~\cite{genm3}               & ICCV 2025 & Multi-path AR  & Multi-expert VQ \\
\midrule
\multicolumn{4}{l}{\textit{Diffusion / Flow Matching}} \\
\quad MDM~\cite{mdm}                   & ICLR 2023 & Diffusion      & -- \\
\quad ViMoGen~\cite{vimogen} & ICLR 2026 & Flow Matching& -- \\
\midrule
\multicolumn{4}{l}{\textit{Improved Tokenization}} \\
\quad DisCoRD~\cite{discord}           & ICCV 2025 & AR + Flow      & VQ + Rect.\ Flow \\
\midrule
\textbf{MotionVLA (Ours)}$^\dagger$ & -- & AR (Qwen LoRA) & \textbf{DS-FAST} \\
\bottomrule
\end{tabular}
\end{table}

\paragraph{Evaluation Metrics.}
On MBench, we follow the official benchmark protocol and report eight dimensions. Motion-Condition Consistency measures whether the generated motion matches the input condition; Motion Generalizability evaluates semantic plausibility and diversity under unseen prompts; Jitter Degree measures local temporal instability; Dynamic Degree evaluates motion expressiveness; Foot Floating and Foot Sliding quantify contact realism; Body Penetration measures self-intersection artifacts; and Pose Quality evaluates overall pose naturalness.

On HumanML3D, we follow the standard text-to-motion evaluation protocol. FID measures the distribution distance between generated and ground-truth motions in the feature space; R-Precision evaluates text-motion retrieval accuracy; MM-Dist measures multimodal alignment distance; Diversity measures sample diversity across generated motions; and MModality evaluates motion variation under the same text condition.

\paragraph{Data Splits and Protocol.}
For ViMoGen-228K, we use the official training split for model training and report results on the MBench evaluation set. For HumanML3D, we follow the official train/test split and the standard evaluation pipeline. All reported numbers are obtained under the corresponding benchmark protocols, and additional implementation details are provided in Appendix~\ref{app:impl}.

\section{Implementation Details}
\label{app:impl}

\paragraph{DS-FAST Tokenizer Training.}
For \textbf{ViMoGen}, the tokenizer is trained on 41{,}971 in-the-wild video motions
and 50{,}000 randomly sampled optical motion-capture sequences from AMASS.
The 276-dim vector is split into Base ($D_b{=}201$) and Phys ($D_p{=}75$) by index slicing.
For \textbf{HumanML3D}, the tokenizer is trained on the 23{,}384 official training sequences;
the 263-dim vector is split into Base ($D_b{=}190$, indices $[7{:}197]$) and
Phys ($D_p{=}73$, indices $[0{:}7]\cup[197{:}263]$).
In both settings, DCT truncation lengths are $K_b{=}5$ and $K_p{=}25$, and two
independent BPE vocabularies of size 4{,}096 are trained per dataset.
The trained tokenizers are then applied to all training samples: 212{,}913 for ViMoGen
(41{,}971 in-the-wild video $+$ 170{,}942 optical mocap) and 23{,}384 for HumanML3D.

\paragraph{Phase 1 — Embedding Cold Start.}
The 8{,}195 newly added motion token embeddings are randomly initialized.
All Qwen3.5 transformer layers are frozen; only \texttt{embed\_tokens} and
\texttt{lm\_head} are trained for 500 steps with learning rate $1{\times}10^{-3}$
and the Adafactor optimizer to warm up the motion token embedding space.

\paragraph{Phase 2 — LoRA Fine-Tuning.}
Starting from the Phase-1 checkpoint, LoRA adapters are applied to all linear
projections, while \texttt{embed\_tokens} and \texttt{lm\_head} continue to be
updated as full saved modules.
Training runs for 10 epochs on $8{\times}$H100 (80\,GB) GPUs.
All base Q wenweights remain frozen throughout.

\paragraph{Training Data.}
ViMoGen mixes \emph{In-the-Wild Video} (41{,}971 samples, image $+$ text) and
\emph{Optical MoCap} (170{,}942 samples, text-only, real GT from AMASS).
HumanML3D uses the official train/val/test split
(23{,}384\,/\,1{,}460\,/\,4{,}384) with text-only inputs.

\paragraph{Inference.}
The model runs on a single H100 (80\,GB) GPU.
The phase-aware logit mask constrains autoregressive decoding to Base tokens
before \texttt{SEP} and Phys tokens after \texttt{SEP}.
Generated tokens are decoded via BPE inverse mapping followed by IDCT to
reconstruct the full motion sequence (276-dim for ViMoGen, 263-dim for HumanML3D).

\paragraph{Hyperparameters.}
Table~\ref{tab:hparams} lists the complete configuration.

\begin{table}[ht]
\centering
\small
\caption{Full hyperparameter configuration for MotionVLA training.}
\label{tab:hparams}
\begin{tabular}{ll}
\toprule
\textbf{Hyperparameter} & \textbf{Value} \\
\midrule
\multicolumn{2}{l}{\textit{DS-FAST Tokenizer}} \\
\quad Base stream dims $D_b$         & 201 (ViMoGen) / 190 (HumanML3D) \\
\quad Phys stream dims $D_p$         & 75 (ViMoGen) / 73 (HumanML3D) \\
\quad DCT truncation $K_b$           & 5 \\
\quad DCT truncation $K_p$           & 25 \\
\quad BPE vocabulary per stream      & 4{,}096 \\
\midrule
\multicolumn{2}{l}{\textit{Vocabulary Expansion}} \\
\quad Qwen3.5 original vocab      & 248{,}320 \\
\quad Total vocab after expansion    & 256{,}515 \\
\quad New motion token embeddings    & 8{,}195 \\
\midrule
\multicolumn{2}{l}{\textit{Phase 1 — Embedding Cold Start}} \\
\quad Trainable parameters           & \texttt{embed\_tokens}, \texttt{lm\_head} \\
\quad Learning rate                  & $1\times10^{-3}$ \\
\quad Optimizer                      & Adafactor \\
\quad Steps                          & 500 \\
\midrule
\multicolumn{2}{l}{\textit{Phase 2 — LoRA Fine-Tuning}} \\
\quad LoRA rank $r$                  & 32 \\
\quad LoRA scaling $\alpha$          & 64 \\
\quad Applied to                     & All linear projections \\
\quad Learning rate                  & $2\times10^{-4}$ \\
\quad LR schedule                    & Cosine decay \\
\quad Optimizer                      & AdamW \\
\quad Epochs                         & 10 \\
\quad Per-device batch size          & 3 \\
\quad Gradient accumulation steps    & 2 \\
\quad Effective batch size           & 48 ($8$ GPUs $\times$ 3 $\times$ 2) \\
\quad Max sequence length            & 4{,}700 tokens \\
\quad Hardware                       & $8\times$ H100 (80\,GB) \\
\midrule
\multicolumn{2}{l}{\textit{Training Data}} \\
\quad ViMoGen total samples          & 212{,}913 \\
\quad \quad In-the-Wild Video        & 41{,}971 (image $+$ text) \\
\quad \quad Optical MoCap            & 170{,}942 (text-only) \\
\quad HumanML3D training samples     & 23{,}384 (text-only) \\
\bottomrule
\end{tabular}
\end{table}

\section{DS-FAST Feature Partition Details}
\label{app:dsfast_partition}

The Base/Phys partition assigns each dimension to the stream whose frequency profile
it matches, determined by the low-frequency energy ratio (LFR) threshold of $0.6$.
Table~\ref{tab:partition} gives the complete per-field mapping for both
ViMoGen (276-dim, SMPL+X) and HumanML3D (263-dim).

\begin{table*}[ht]
\centering
\small
\setlength{\tabcolsep}{5pt}
\caption{%
  Per-field breakdown of the ViMoGen 276-dim and HumanML3D 263-dim motion vectors
  into Base ($D_b$, position/rotation) and Phys ($D_p$, velocity) streams.
  Index ranges index into the flat per-frame feature vector $\mathbf{d}$.}
\label{tab:partition}
\begin{tabular}{llcccc}
\toprule
\textbf{Field} & \textbf{Semantics} & \textbf{Dims} & \textbf{Index range} & \textbf{Stream} \\
\midrule
\multicolumn{5}{l}{%
  \textbf{ViMoGen} \; (276 dims, \;
  Base $D_b{=}201$, \; Phys $D_p{=}75$)} \\[3pt]
\multicolumn{5}{l}{\quad\textit{Base stream — position and rotation}} \\
\quad \texttt{body\_pose\_6d}   & 21 joints $\times$ 6D rotation         & 126 & \texttt{[:,\;0:126]}   & Base \\
\quad \texttt{joints}           & 22 joints $\times$ XYZ position        &  66 & \texttt{[:,\;126:192]} & Base \\
\quad \texttt{root\_orient\_6d} & Root global orientation (6D)           &   6 & \texttt{[:,\;258:264]} & Base \\
\quad \texttt{root\_trans}      & Root global translation (XYZ)          &   3 & \texttt{[:,\;270:273]} & Base \\[2pt]
\multicolumn{5}{l}{\quad\textit{Phys stream — velocity}} \\
\quad \texttt{joints\_vel}      & 22 joints $\times$ XYZ velocity        &  66 & \texttt{[:,\;192:258]} & Phys \\
\quad \texttt{root\_vel\_6d}    & Root rotational velocity (6D)          &   6 & \texttt{[:,\;264:270]} & Phys \\
\quad \texttt{root\_trans\_vel} & Root translational velocity (XYZ)      &   3 & \texttt{[:,\;273:276]} & Phys \\
\midrule
\multicolumn{5}{l}{%
  \textbf{HumanML3D} \; (263 dims, \;
  Base $D_b{=}190$, \; Phys $D_p{=}73$)} \\[3pt]
\multicolumn{5}{l}{\quad\textit{Base stream — position and rotation}} \\
\quad \texttt{local\_pos}       & Non-root joints XYZ position ($20{\times}3$) & 60  & \texttt{[:,\;7:67]}    & Base \\
\quad \texttt{local\_rot}       & 21 joints $\times$ 6D rotation               & 126 & \texttt{[:,\;67:193]}  & Base \\
\quad \texttt{root\_joint\_vel} & Root joint velocity (low LFR)                &   4 & \texttt{[:,\;193:197]} & Base \\[2pt]
\multicolumn{5}{l}{\quad\textit{Phys stream — velocity and root dynamics}} \\
\quad \texttt{root\_ang\_vel}   & Root angular velocity (Y-axis)               &   1 & \texttt{[:,\;0:1]}     & Phys \\
\quad \texttt{root\_lin\_vel}   & Root linear velocity (X, Z)                  &   2 & \texttt{[:,\;1:3]}     & Phys \\
\quad \texttt{root\_height}     & Root height (Y)                              &   1 & \texttt{[:,\;3:4]}     & Phys \\
\quad \texttt{root\_pos}        & Root joint XYZ position                      &   3 & \texttt{[:,\;4:7]}     & Phys \\
\quad \texttt{local\_vel}       & Non-root joint velocities ($62$ dims)        &  62 & \texttt{[:,\;197:259]} & Phys \\
\quad \texttt{foot\_contact}    & Foot contact binary labels                   &   4 & \texttt{[:,\;259:263]} & Phys \\
\midrule
\multicolumn{2}{l}{\textbf{Total — ViMoGen}}   & \textbf{276} & \texttt{[:,\;0:276]} & --- \\
\multicolumn{2}{l}{\textbf{Total — HumanML3D}} & \textbf{263} & \texttt{[:,\;0:263]} & --- \\
\bottomrule
\end{tabular}
\end{table*}

In the ViMoGen representation, velocity fields occupy non-contiguous index ranges:
\texttt{joints\_vel} ($[192{:}258]$) is interleaved between the Base \texttt{joints}
block ($[126{:}192]$) and the Base \texttt{root\_orient\_6d} block ($[258{:}264]$).
DS-FAST extracts both streams by explicit index slicing prior to DCT, ensuring
that each stream contains only physically homogeneous features.

In HumanML3D, the LFR boundary falls within two semantic fields rather than
between them: the 21-joint \texttt{local\_pos} block is split so that the root
joint position ($[4{:}7]$) enters the Phys stream due to its high-frequency root
dynamics, while the remaining non-root positions ($[7{:}67]$) remain in Base;
similarly, the first four elements of the 66-dim joint velocity block ($[193{:}197]$),
which correspond to the root joint and exhibit low LFR, are assigned to the Base stream.
These splits reflect the data-driven LFR criterion and do not require any manual
field-boundary annotation.

\section{Human Preference Analysis}
\label{sec:human_study}

\begin{figure}[t]
  \centering
  \includegraphics[width=\linewidth]{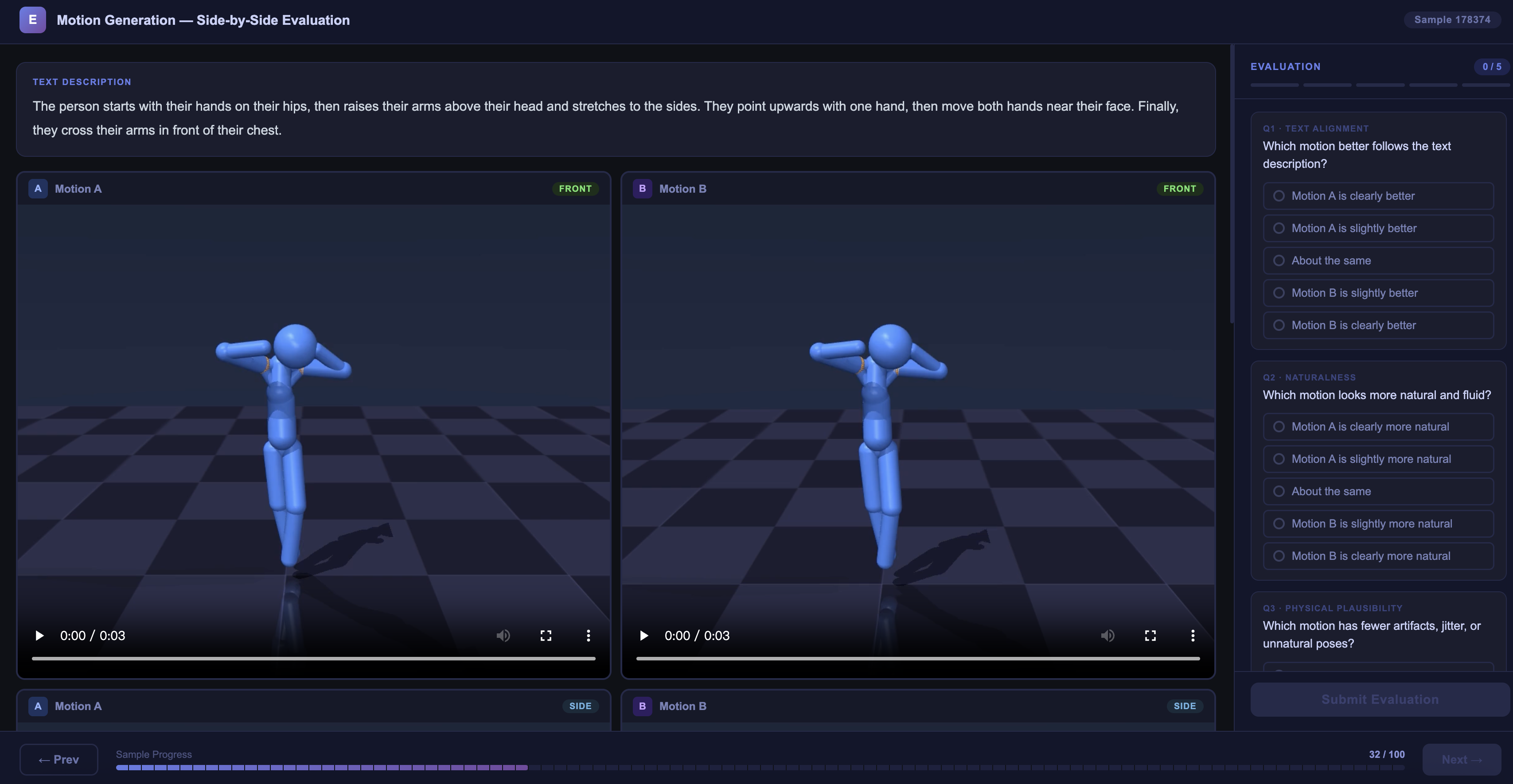}
  \caption{
    Screenshot of the GSB evaluation interface used in our human preference study.
    Each trial displays two motion clips—Motion~A and Motion~B—rendered from
    \textit{front} and \textit{side} viewpoints simultaneously, with the
    conditioning text prompt shown above.
    Experts select one of three options: \textbf{G} (left clip better),
    \textbf{S} (comparable quality), or \textbf{B} (right clip better).
    The right panel tracks per-question completion progress across
    five evaluation dimensions.
  }
  \label{fig:gsb_interface}
\end{figure}

To complement the quantitative benchmarks reported in
Section~\ref{sec:experiments}, we conducted a human preference study
to assess the perceptual quality of motions generated by MotionVLA.
Evaluations were carried out through a custom web-based interface
(Figure~\ref{fig:gsb_interface}) that presented anonymized side-by-side
motion pairs to domain experts.

\paragraph{Study Design.}
We invited five domain experts in human motion analysis and character animation
to participate in a blinded pairwise preference evaluation.
Each expert assessed \textbf{100 text-conditioned motion pairs}, where each pair
comprised one motion generated by MotionVLA and the corresponding output from
ViMoGen~\cite{vimogen}.
To enable a comprehensive visual assessment, every motion clip was rendered from
two camera perspectives—\textit{front view} and \textit{side view}—yielding
four synchronized video clips per evaluation trial.
Each clip was rendered as a 3-second video at 20\,fps with the conditioning text
prompt displayed above both clips.
The left/right assignment of MotionVLA and the baseline was randomized per trial;
experts were not informed of which method produced either clip.
All participants volunteered without monetary compensation.
As the study involved only viewing and comparing AI-generated skeletal motion
clips with no collection of personal data, IRB approval was not required under
our institutional guidelines.

\paragraph{Evaluation Protocol.}
For each pair, the expert selected one of three options:

\medskip
\noindent
\textbf{Good (G):} The left clip is clearly better overall.\quad
\textbf{Same (S):} The two clips are of comparable quality.\quad
\textbf{Bad (B):} The right clip is clearly better overall.
\medskip

\noindent
After de-anonymizing, \textbf{G} indicates a preference for MotionVLA,
\textbf{S} indicates no clear preference, and \textbf{B} indicates a preference
for the baseline.
Preference rates (\%) are reported over all
$5\text{ experts} \times 100\text{ prompts} = 500$ comparisons.

\paragraph{Aggregate Results.}
Table~\ref{tab:user_study} reports the GSB preference rates of MotionVLA
against ViMoGen, aggregated across all 500 comparisons.
MotionVLA receives a clear majority preference (G\,=\,64.0\%), while only
14.0\% of evaluations favor the baseline, demonstrating a substantial and
consistent advantage in perceived motion quality across both front and side views.

\begin{table}[h]
\centering
\small
\caption{GSB pairwise preference study results (\%).
         G\,=\,MotionVLA preferred; S\,=\,no preference; B\,=\,baseline preferred.
         Results aggregated over 5 experts $\times$ 100 prompts $=$ 500 comparisons.}
\label{tab:user_study}
\begin{tabular}{lccc}
\toprule
\textbf{MotionVLA vs.} & \textbf{G (Ours Better)} & \textbf{S (Same)} & \textbf{B (Baseline Better)} \\
\midrule
ViMoGen~\cite{vimogen} & 64.0 & 22.0 & 14.0 \\
\bottomrule
\end{tabular}
\end{table}

\begin{figure}[t]
  \centering
  \includegraphics[width=\linewidth]{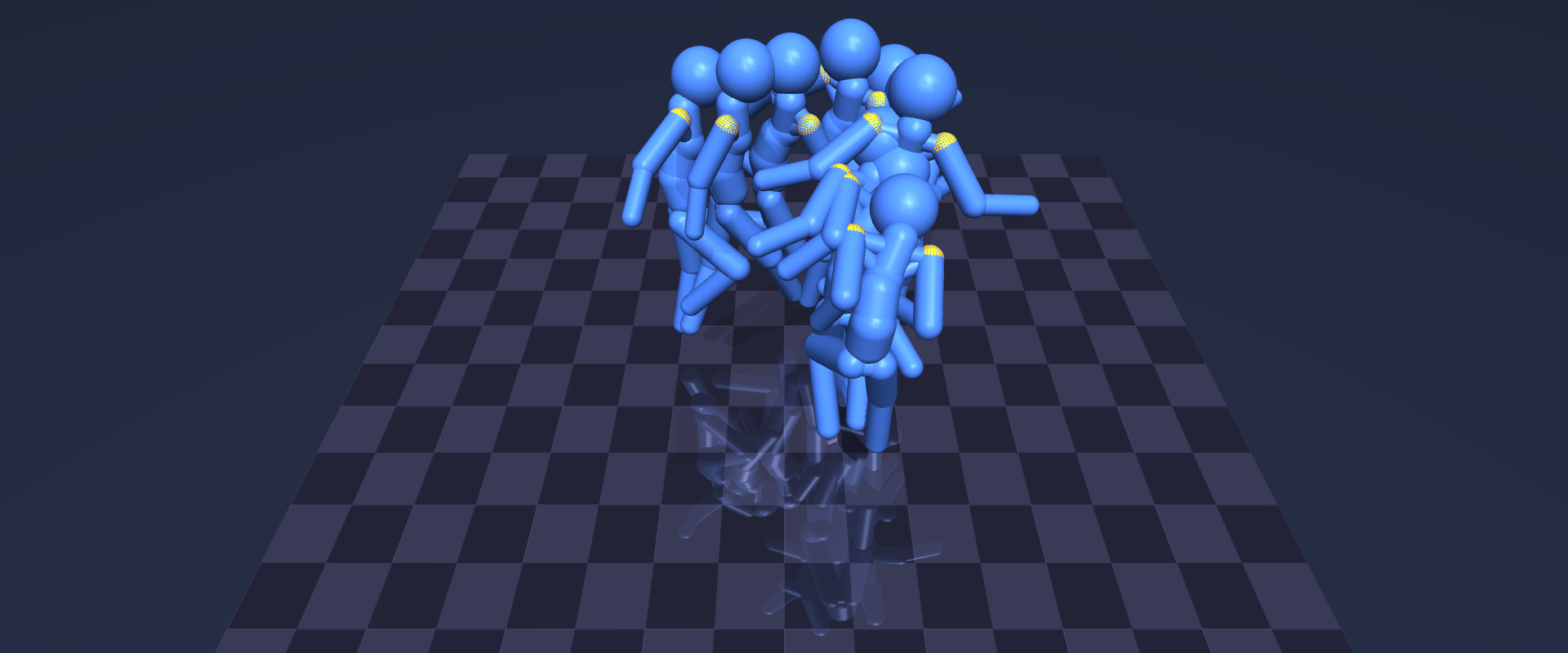}
  \caption{
    Capsule-skeleton rendering produced by MuJoCo. The motion decoded from
    DS-FAST tokens is converted to SMPL-X joint positions and rendered with
    per-sequence ground-contact alignment.
  }
  \label{fig:simulation}
\end{figure}

\section{Simulation and Real-Robot Demonstration}
\label{app:simulation}

\subsection{MuJoCo Simulation}

All qualitative motion visualizations in this paper are produced using
MuJoCo~\cite{mujoco}, a physics engine widely used in locomotion and
character animation research.

\paragraph{Pipeline.}
The generated motion token sequence is first decoded by DS-FAST into a per-frame
motion vector (276-dim for ViMoGen, 263-dim for HumanML3D) through inverse BPE
followed by inverse DCT. This vector is then converted to SMPL-X~\cite{smplx}
body parameters (global orientation, 22-joint body pose, and root translation).
In MuJoCo, each frame is visualized as a capsule-based skeleton, where bone
segments connecting adjacent joints are drawn as capsule geometries and joint
centers are marked by spheres. The scene is evaluated with \texttt{mj\_forward}
in pure kinematic mode (no physical integration), so the rendered motion exactly
reflects the model output without any simulation correction.

\paragraph{Rendering Configuration.}
Frames are rendered with the MuJoCo offscreen renderer (EGL backend) at
$1280{\times}1024$ resolution and composited at 20\,fps under a fixed side-view
camera. To ensure plausible ground contact, a per-sequence vertical offset aligns
the lowest foot position with the floor plane.

\subsection{Real-Robot Deployment}

We further deploy MotionVLA on a Unitree G1 EDU humanoid robot to verify that the
generated motions can be executed on real hardware. Given a text prompt, the
model produces a motion token sequence, which DS-FAST decodes into joint-angle
trajectories. These trajectories are retargeted to the G1 joint configuration
and executed in real time.

\begin{figure}[t]
  \centering
  \begin{minipage}{0.32\linewidth}\centering
    \includegraphics[width=\linewidth]{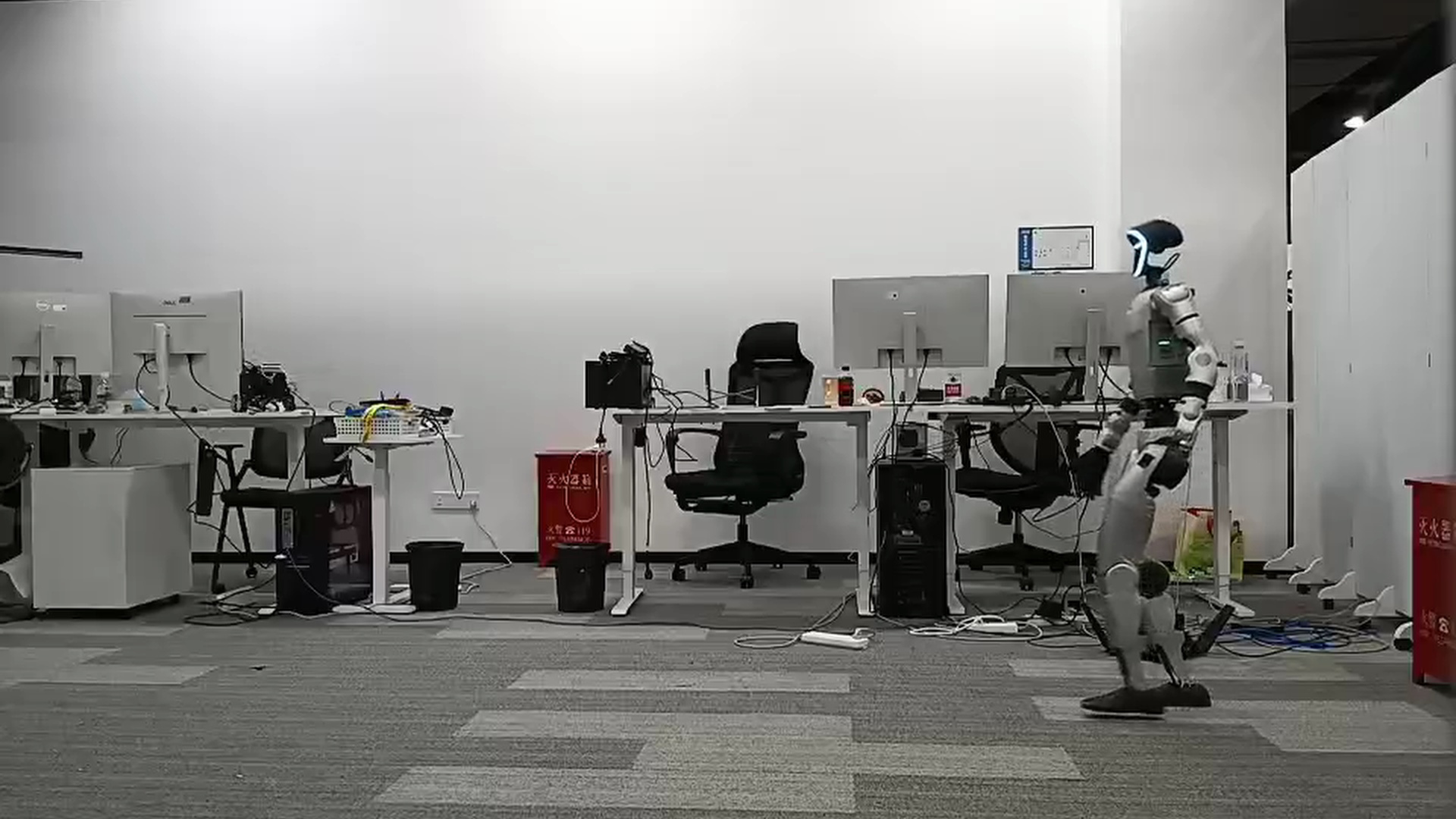}
  \end{minipage}\hfill
  \begin{minipage}{0.32\linewidth}\centering
    \includegraphics[width=\linewidth]{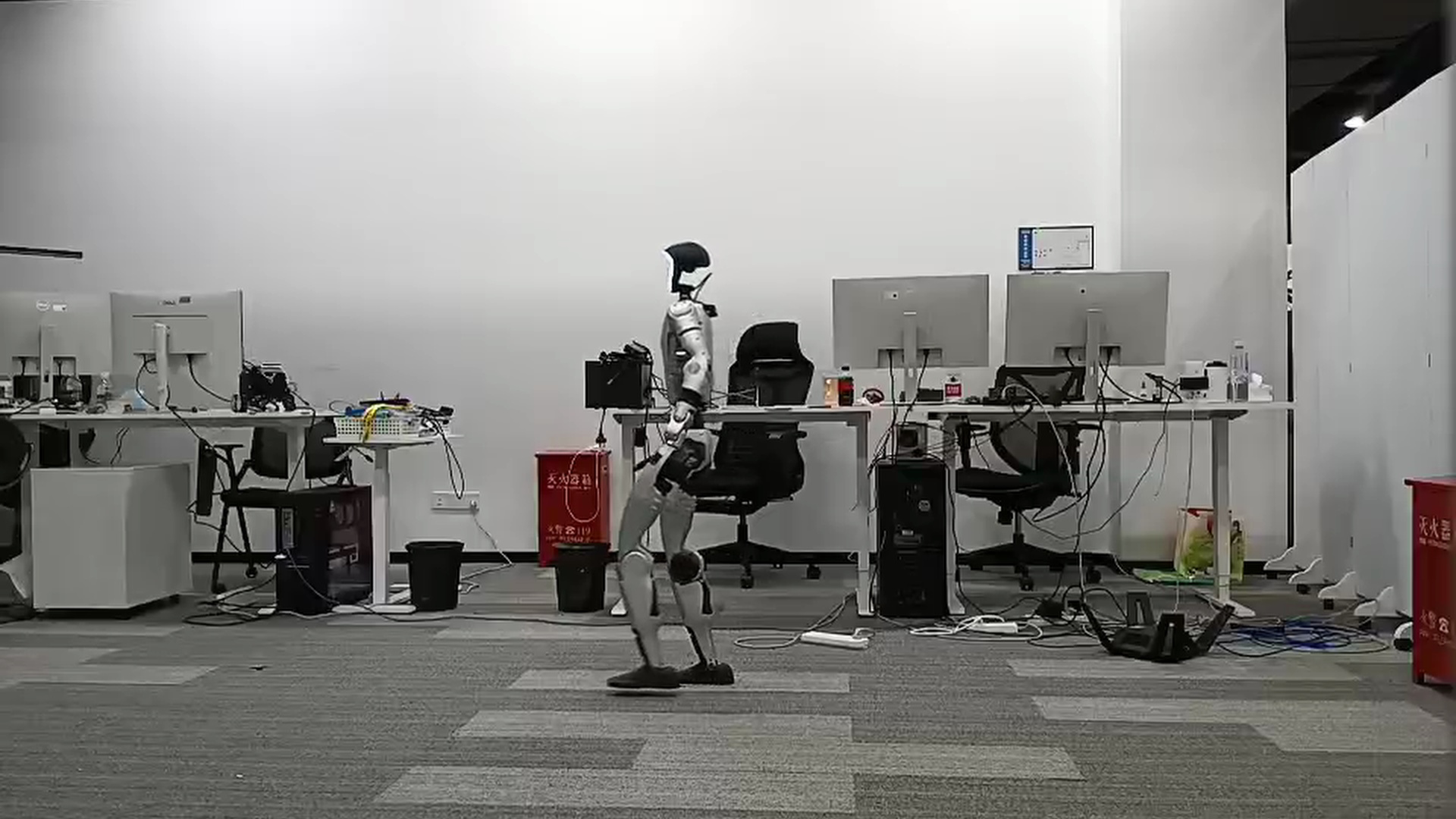}
  \end{minipage}\hfill
  \begin{minipage}{0.32\linewidth}\centering
    \includegraphics[width=\linewidth]{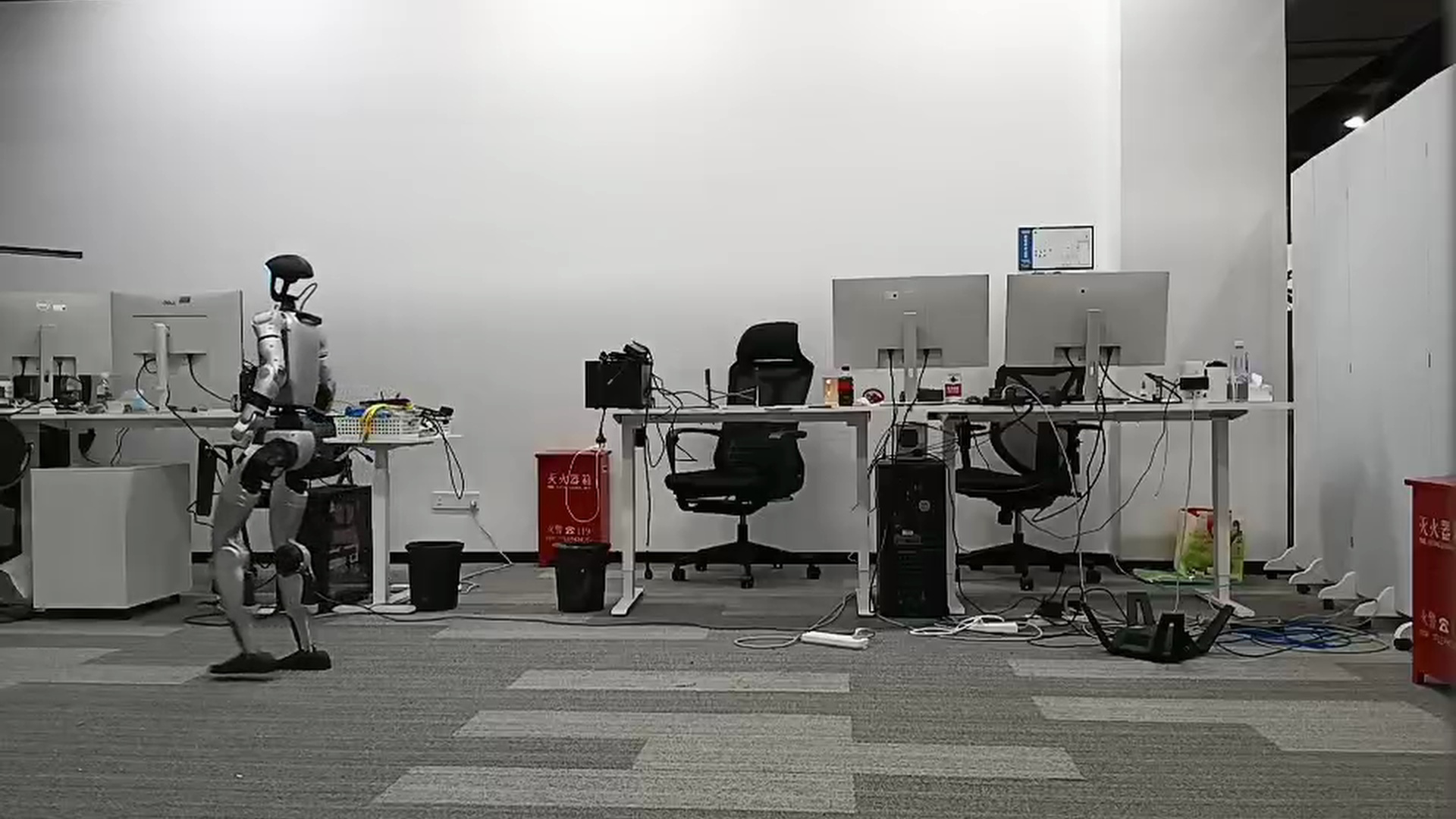}
  \end{minipage}\\[2pt]
  {\small (a) The person walks straight ahead to the other end of the room.}

  \vspace{6pt}
  \begin{minipage}{0.32\linewidth}\centering
    \includegraphics[width=\linewidth]{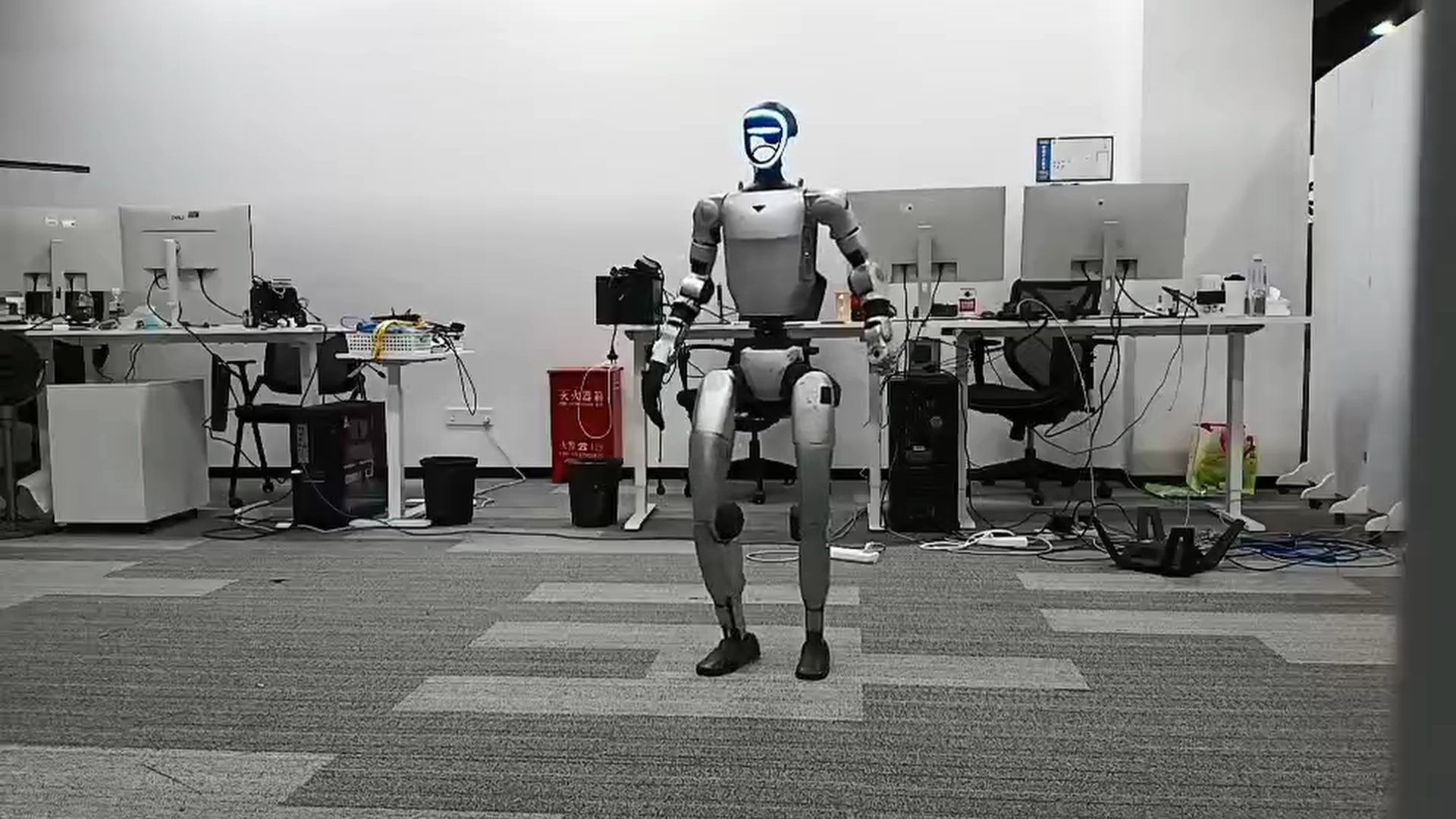}
  \end{minipage}\hfill
  \begin{minipage}{0.32\linewidth}\centering
    \includegraphics[width=\linewidth]{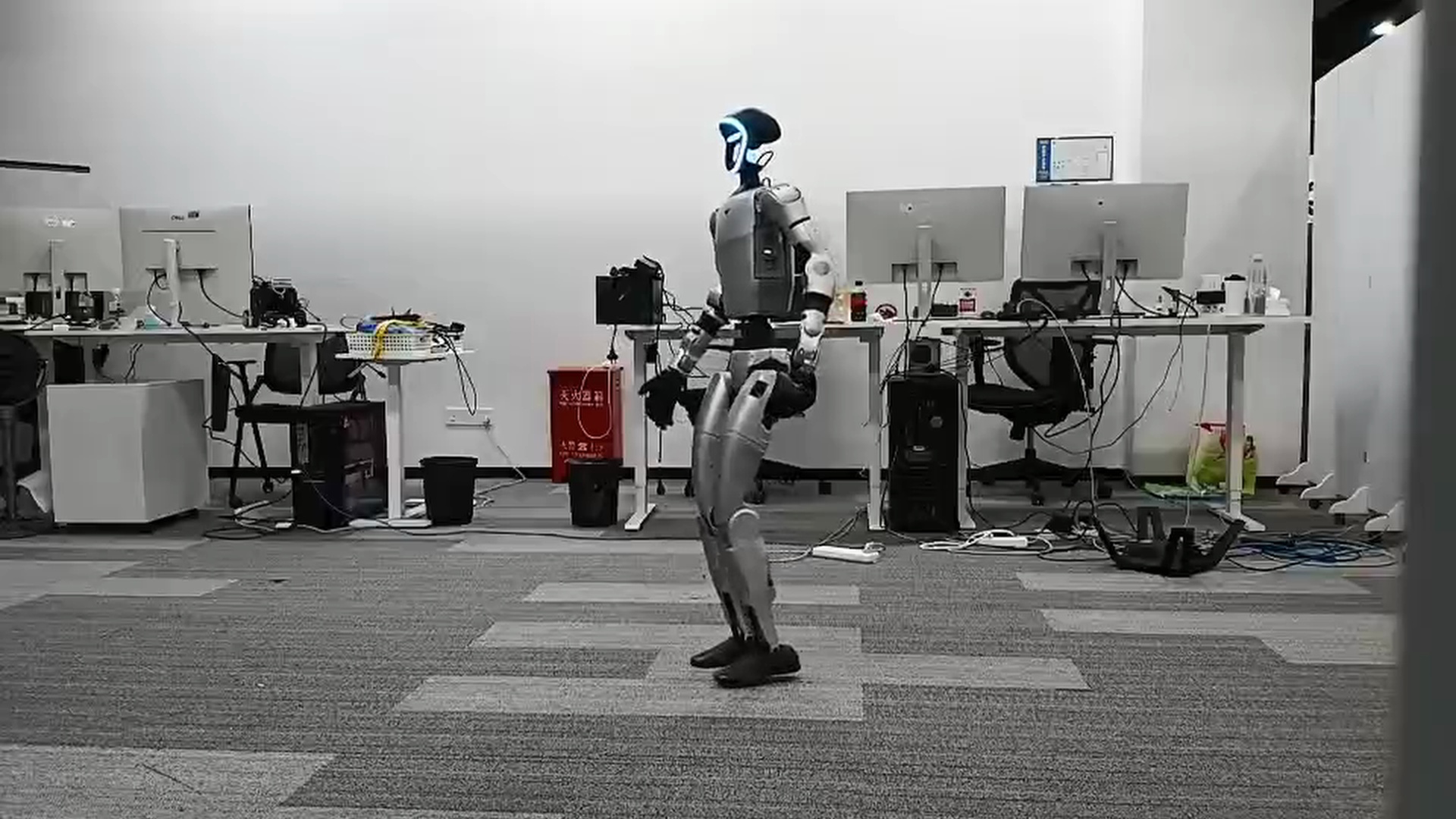}
  \end{minipage}\hfill
  \begin{minipage}{0.32\linewidth}\centering
    \includegraphics[width=\linewidth]{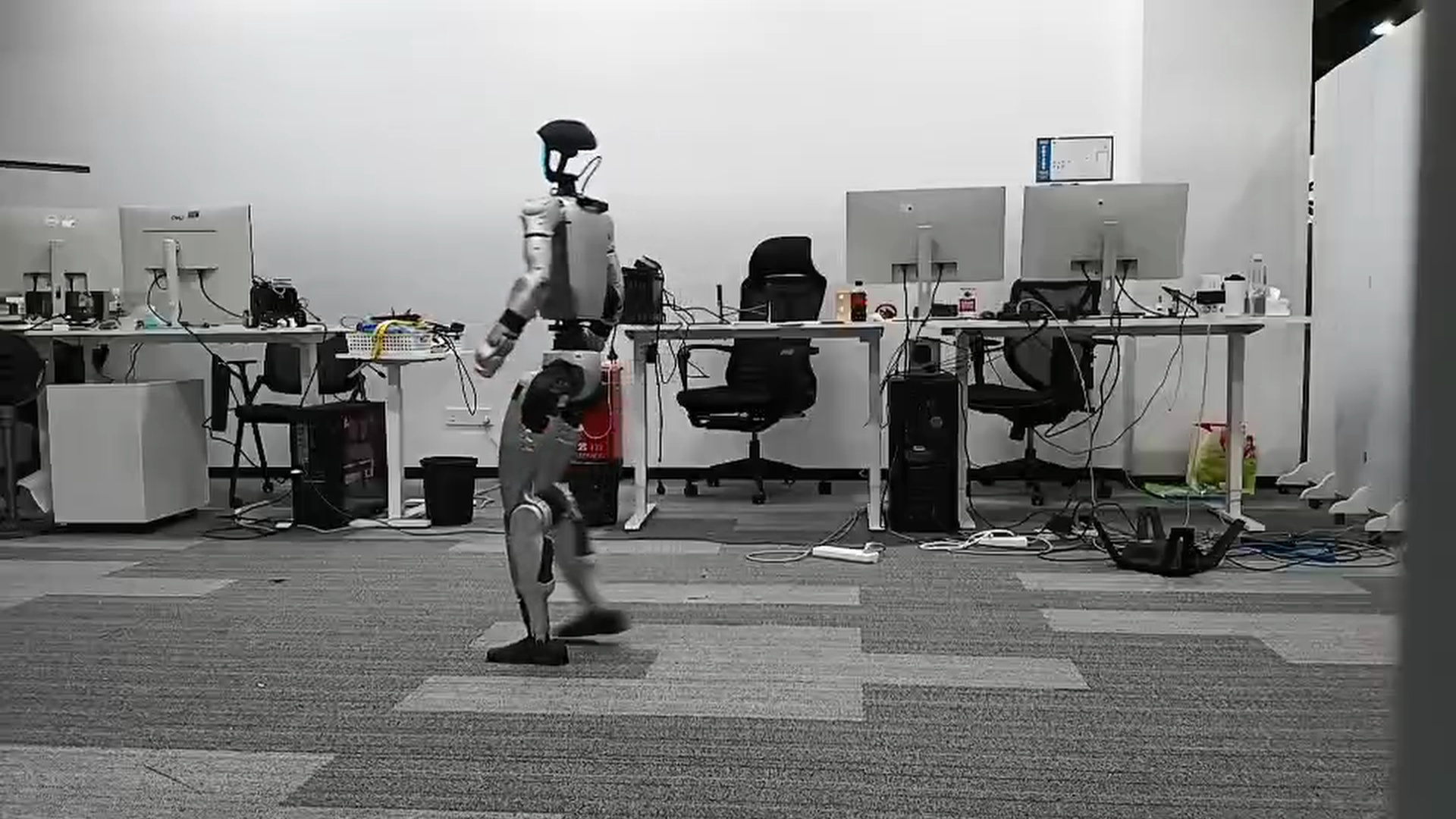}
  \end{minipage}\\[2pt]
  {\small (b) The person turns and then walks to the end of the room.}

  \vspace{6pt}
  \begin{minipage}{0.32\linewidth}\centering
    \includegraphics[width=\linewidth]{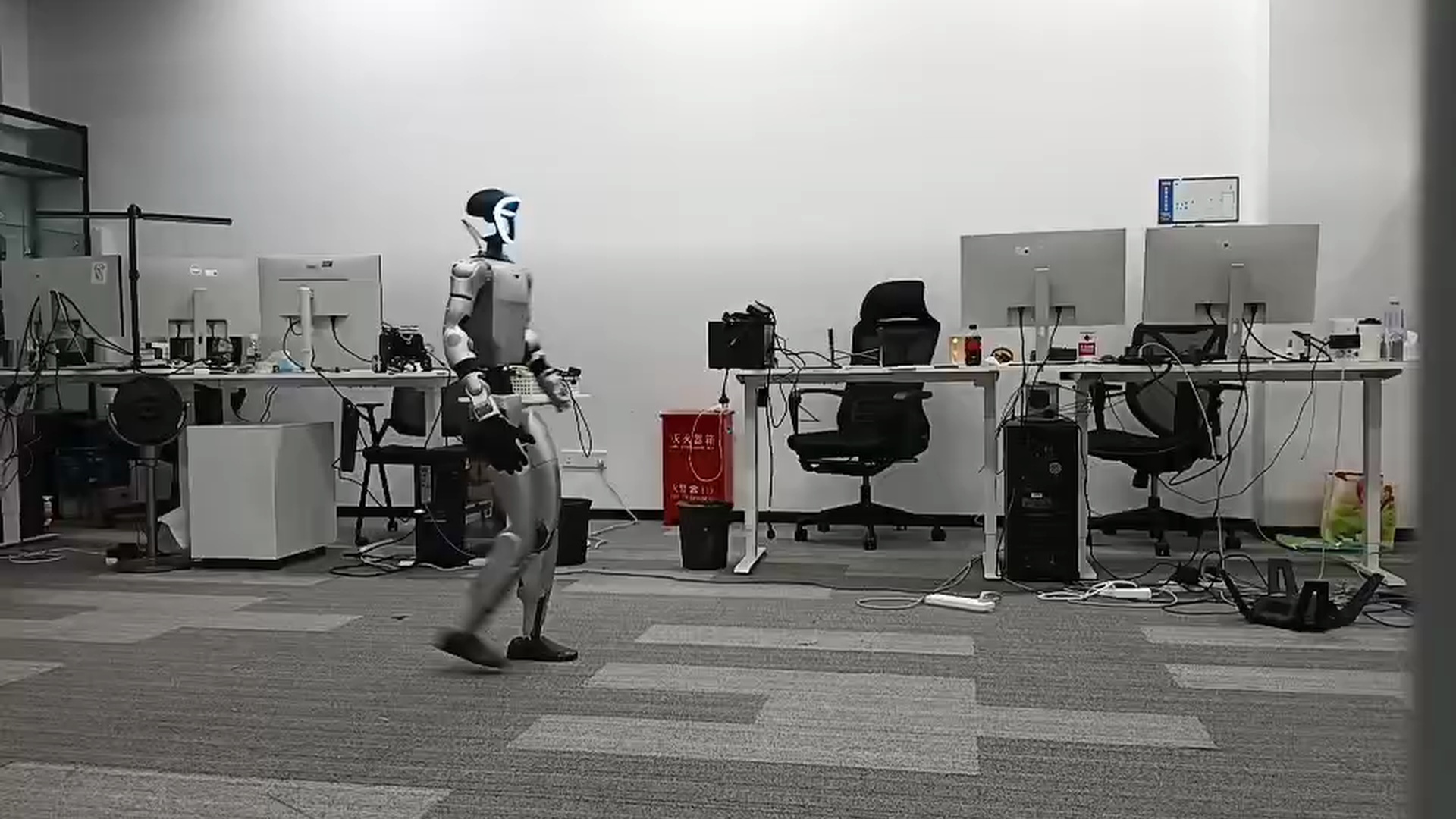}
  \end{minipage}\hfill
  \begin{minipage}{0.32\linewidth}\centering
    \includegraphics[width=\linewidth]{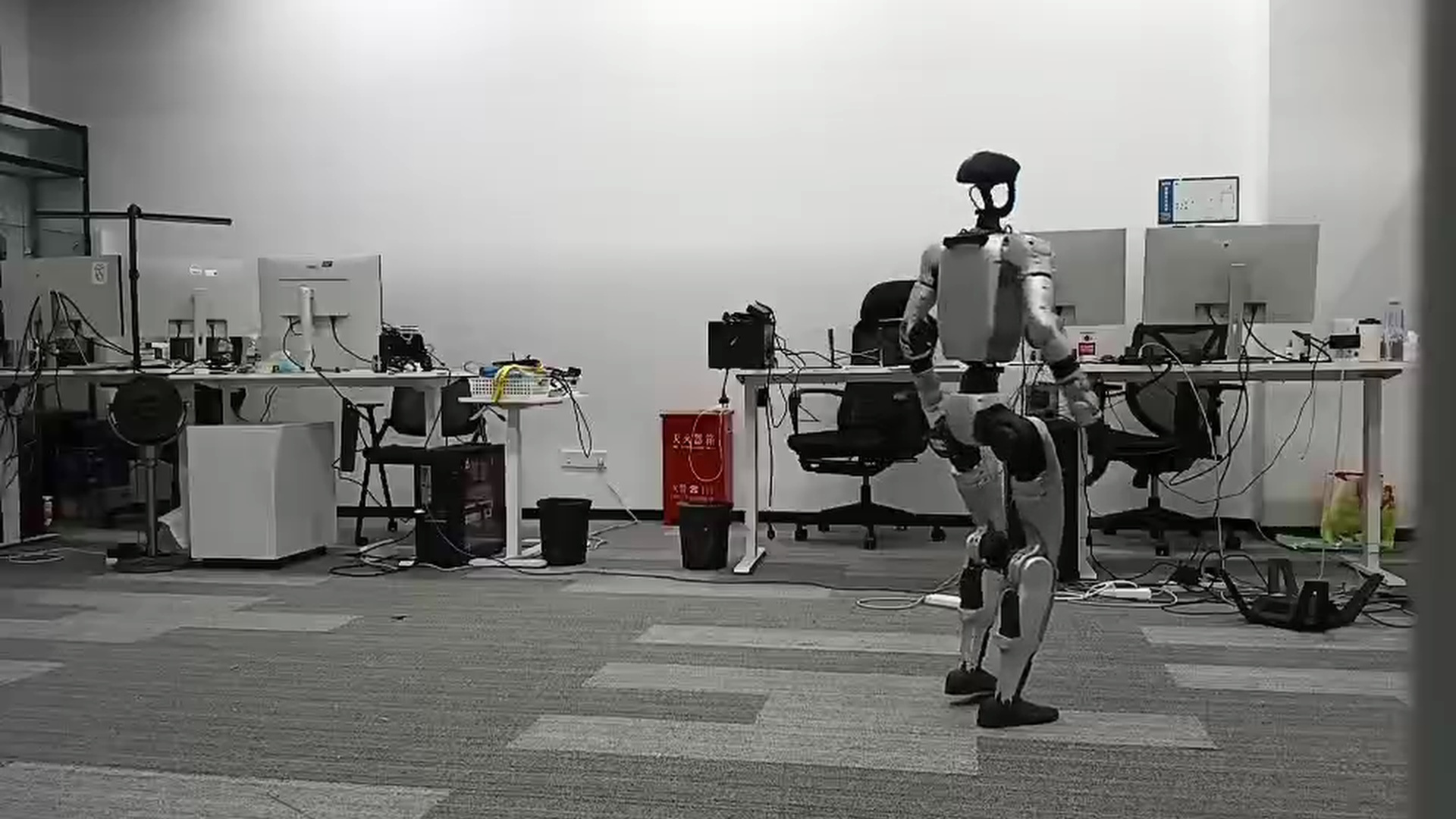}
  \end{minipage}\hfill
  \begin{minipage}{0.32\linewidth}\centering
    \includegraphics[width=\linewidth]{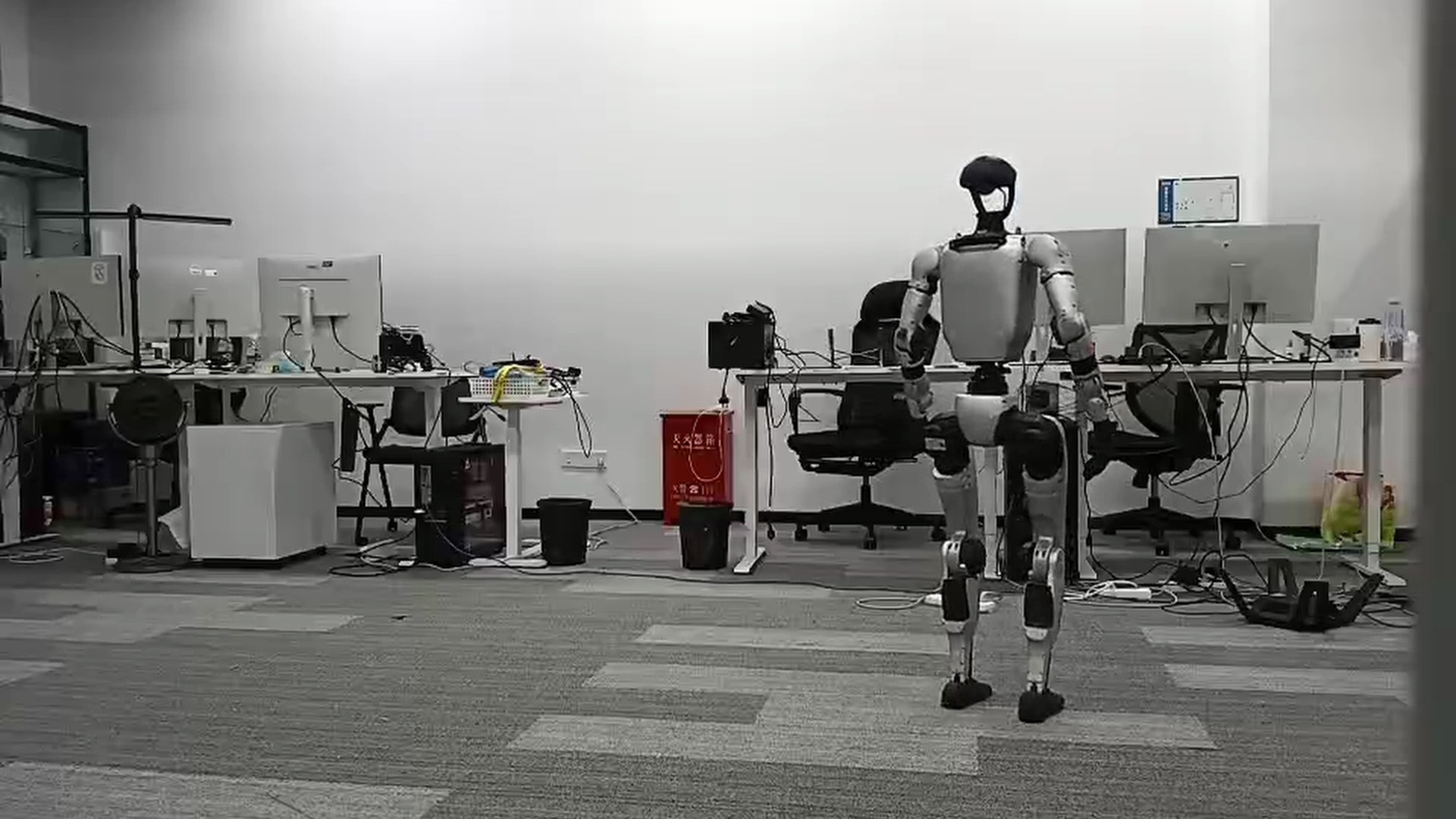}
  \end{minipage}\\[2pt]
  {\small (c) The person walks straight ahead and then turns.}

  \caption{
    Real-robot deployment of MotionVLA on a Unitree G1 EDU humanoid robot.
    Each row shows three exocentric frames from one text-conditioned motion
    execution, captured at different time steps.
  }
  \label{fig:real_robot}
\end{figure}

Figure~\ref{fig:real_robot} presents three deployment examples. Each row
corresponds to one text prompt, showing three exocentric frames captured at
successive time steps.

\newtcolorbox{casestudy}[2][]{
  enhanced,
  breakable,
  colback=gray!10!white,
  colframe=blue!75!black,
  colbacktitle=blue!75!black,
  coltitle=white,
  fonttitle=\bfseries\large,
  title={#2},
  rounded corners,
  arc=3mm,
  boxrule=2pt,
  left=10pt, right=10pt,
  top=5pt, bottom=10pt,
  toptitle=2mm, bottomtitle=2mm,
  before skip=6pt,
  after skip=10pt,
  break at=0pt,
  pad at break=1mm,
  #1
}
\section{Case Study: Scene-Conditioned Motion Generation}
\label{app:case_study}

\vspace{-0.5em}

\begin{casestudy}{End-to-End Generation Example \#1}

\textbf{Text prompt}\\[4pt]
\textit{``Generate motion for: The person takes off a shirt and puts it on
their head, then bends down to pick up something from the ground.''}

\medskip
\begin{center}
\includegraphics[width=0.55\linewidth]{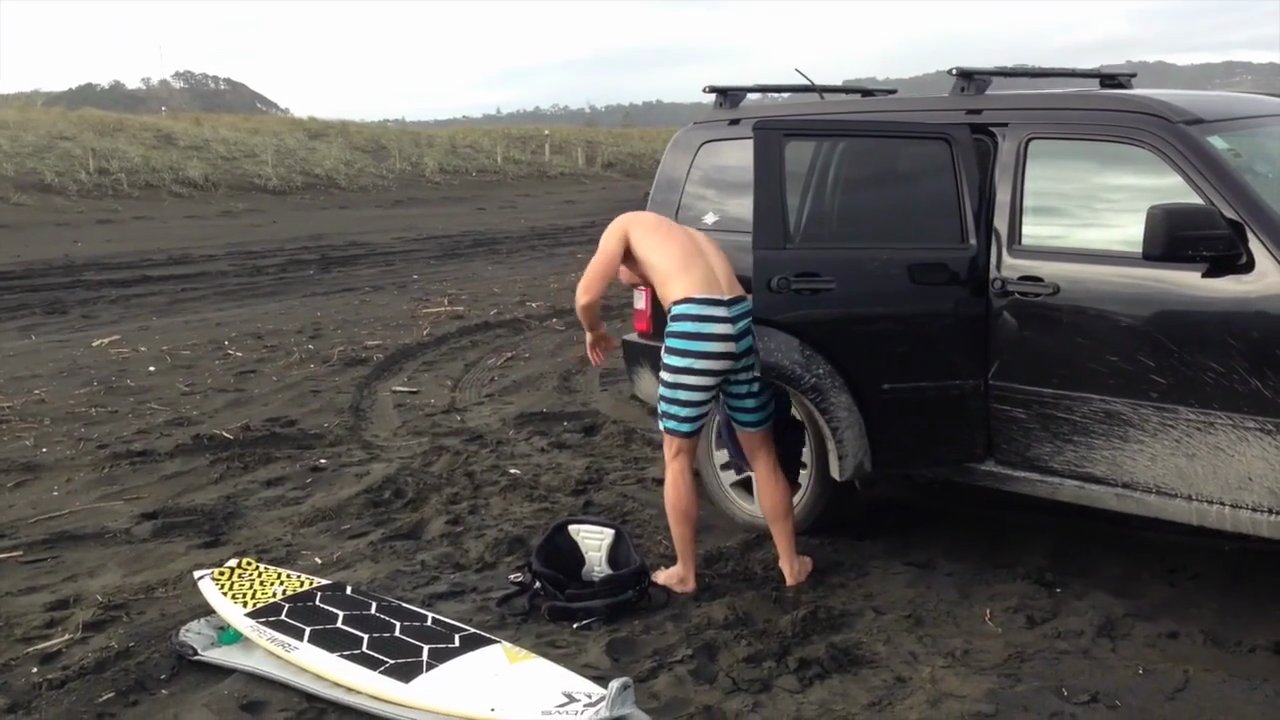}
\\[3pt]
\small\textit{Scene image input}
\end{center}
\raggedright

\medskip\hrule\medskip

\textbf{Generated token sequence.}
Structural markers are highlighted:
\colorbox{teal!20}{\texttt{<mot\_bos>}} opens the sequence,
\colorbox{yellow!50}{\texttt{<mot\_sep>}} separates Base and Phys streams,
\colorbox{red!20}{\texttt{<mot\_eos>}} closes it.
{\color{blue}Blue} = Base stream ($b_i$);
{\color{orange!85!black}orange} = Phys stream ($p_j$).

\smallskip
\begin{tcolorbox}[
  colback=white, colframe=black!20,
  boxrule=0.6pt, arc=2mm,
  left=6pt, right=6pt, top=5pt, bottom=5pt,
  breakable
]
\small\ttfamily\raggedright\setlength{\baselineskip}{1.6em}%
\colorbox{teal!20}{\textbf{<mot\_bos>}}~%
{\color{blue}%
<mot\_b\_1496> <mot\_b\_2644> <mot\_b\_1521> <mot\_b\_1410>
<mot\_b\_1600> <mot\_b\_1506> <mot\_b\_1477> <mot\_b\_1457>
<mot\_b\_1515> <mot\_b\_1477> <mot\_b\_1528> <mot\_b\_1494>
<mot\_b\_1474> <mot\_b\_1547> <mot\_b\_1422> <mot\_b\_1585>
<mot\_b\_1494> <mot\_b\_1624> <mot\_b\_1506> }~$\cdots$~%
\colorbox{yellow!50}{\textbf{<mot\_sep>}}~%
{\color{orange!85!black}%
<mot\_p\_0298> <mot\_p\_0908> <mot\_p\_0318> <mot\_p\_0413>
<mot\_p\_0431> <mot\_p\_0299> <mot\_p\_0336> <mot\_p\_0458>
<mot\_p\_0240> <mot\_p\_2556> <mot\_p\_2761> <mot\_p\_3571>}~$\cdots$~%
\colorbox{red!20}{\textbf{<mot\_eos>}}%
\end{tcolorbox}

\medskip\hrule\medskip

\textbf{Reconstructed motion.}
Decoded by DS-FAST (inverse BPE $\to$ inverse DCT per stream $\to$ concatenate)
from the token sequence above.

\medskip
\centering
\includegraphics[width=\linewidth]{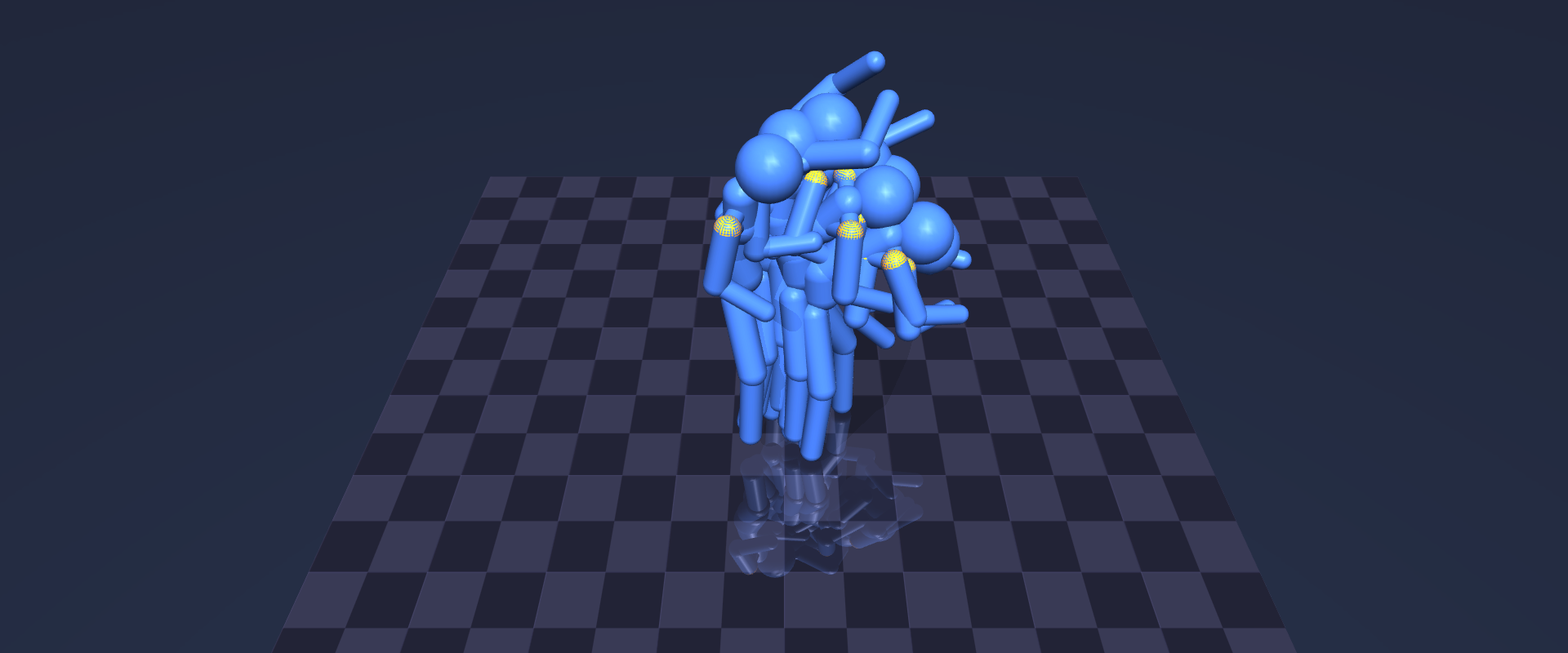}
\\[3pt]
\small\textit{Reconstructed motion sequence visualized at uniform time intervals}

\end{casestudy}

\bigskip

\begin{casestudy}{End-to-End Generation Example \#2}

\textbf{Text prompt}\\[4pt]
\textit{``Generate motion for: The man walks towards the camera.''}

\medskip
\begin{center}
\includegraphics[width=0.55\linewidth]{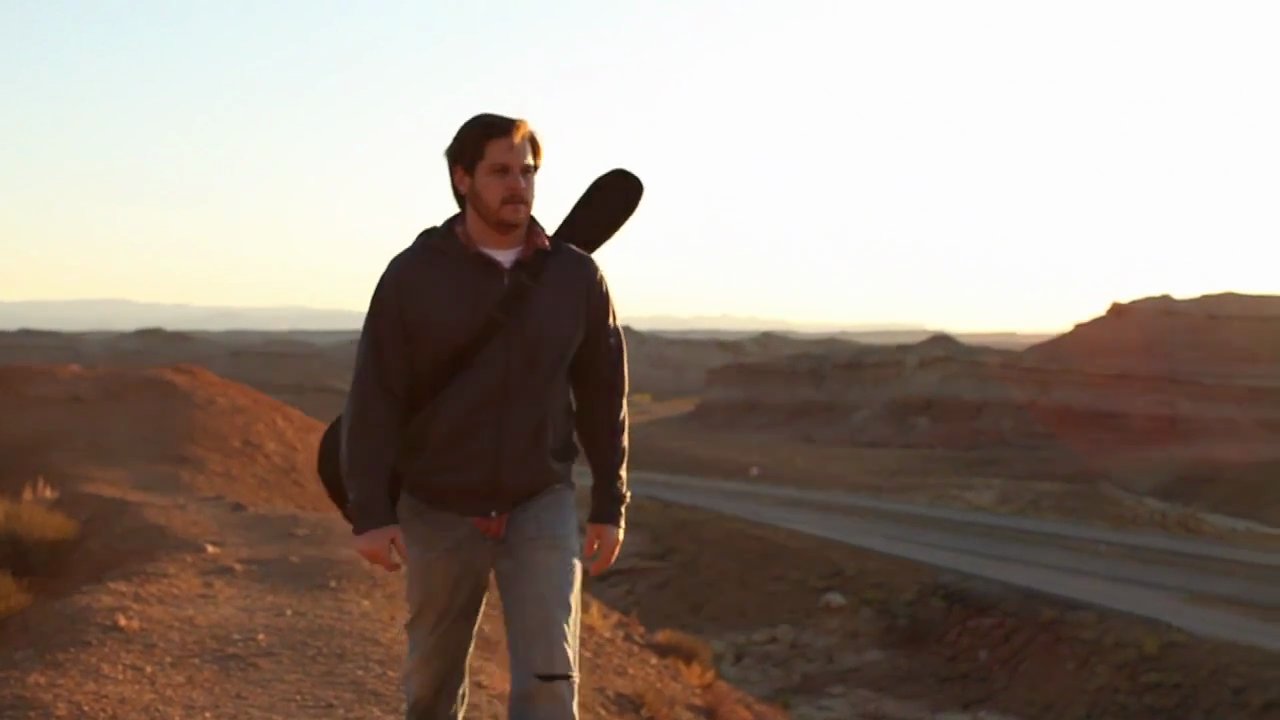}
\\[3pt]
\small\textit{Scene image input}
\end{center}

\medskip\hrule\medskip

\textbf{Generated token sequence.}
Structural markers are highlighted:
\colorbox{teal!20}{\texttt{<mot\_bos>}} opens the sequence,
\colorbox{yellow!50}{\texttt{<mot\_sep>}} separates Base and Phys streams,
\colorbox{red!20}{\texttt{<mot\_eos>}} closes it.
{\color{blue}Blue} = Base stream ($b_i$);
{\color{orange!85!black}orange} = Phys stream ($p_j$).

\smallskip
\begin{tcolorbox}[
  colback=white, colframe=black!20,
  boxrule=0.6pt, arc=2mm,
  left=6pt, right=6pt, top=5pt, bottom=5pt,
  breakable
]
\small\ttfamily\raggedright\setlength{\baselineskip}{1.6em}%
\colorbox{teal!20}{\textbf{<mot\_bos>}}~%
{\color{blue}%
<mot\_b\_1802> <mot\_b\_1570> <mot\_b\_1783> <mot\_b\_2401>
<mot\_b\_1802> <mot\_b\_1416> <mot\_b\_1735> <mot\_b\_1423>
<mot\_b\_1549> <mot\_b\_1802> <mot\_b\_1530> <mot\_b\_1796>
<mot\_b\_1413> <mot\_b\_1483> <mot\_b\_1795> <mot\_b\_2200>
<mot\_b\_1754> <mot\_b\_1410> <mot\_b\_1543>}~$\cdots$~%
\colorbox{yellow!50}{\textbf{<mot\_sep>}}~%
{\color{orange!85!black}%
<mot\_p\_0243> <mot\_p\_0247> <mot\_p\_1751> <mot\_p\_1152>
<mot\_p\_0857> <mot\_p\_1152> <mot\_p\_0857> <mot\_p\_0493>
<mot\_p\_1152> <mot\_p\_0857> <mot\_p\_0493>}~$\cdots$~%
\colorbox{red!20}{\textbf{<mot\_eos>}}%
\end{tcolorbox}

\medskip\hrule\medskip

\textbf{Reconstructed motion.}
Decoded by DS-FAST (inverse BPE $\to$ inverse DCT per stream $\to$ concatenate)
from the token sequence above.

\medskip
\centering
\includegraphics[width=\linewidth]{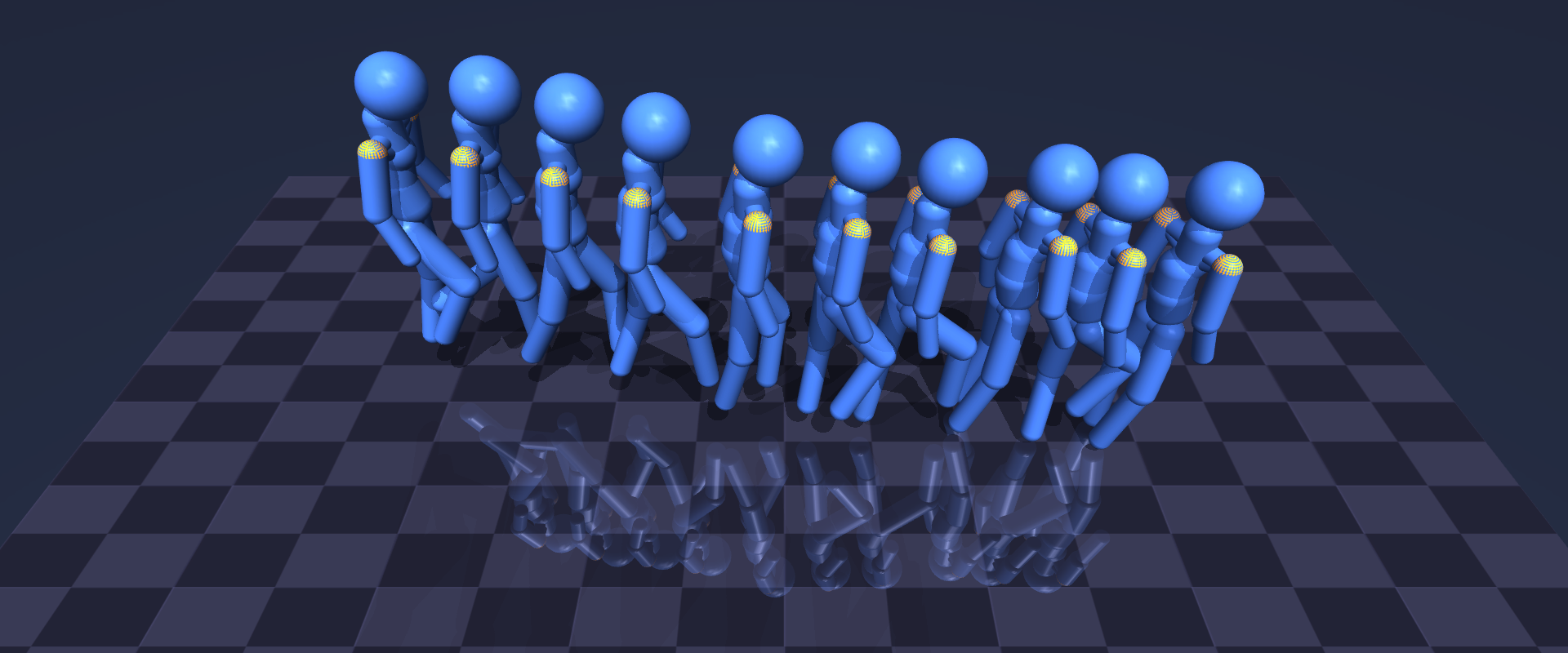}
\\[3pt]
\small\textit{Reconstructed motion sequence visualized at uniform time intervals}

\end{casestudy}

\end{document}